%% file: paper.tex
\documentclass[english]{article}

\input{packages}
\usepackage{bbm}
\usepackage{adjustbox}
\usepackage{amsmath}
\usepackage{placeins}

\input{commands}

\input{title}

\begin{document}


\maketitle

\input{00-Abstract}


\vspace{20pt}

\section{Introduction}
\label{sec:introduction}
\input{01-intro}


\section{Notations and Problem Formulation}
\label{sec:problem-formulation}
\input{02-ProblemFormulation}

\section{Approach  overview}
\label{sec:Approach}
\input{03-Approach}

\section{Single Triplet}
\label{sec:ApproachSingleTriplet}
\input{04-approach_single_triplet}

\section{Data Propagation in a Qualitative Map}
\label{sec:ApproachComposition}
\input{05-approach_composition}

\section{Results}
\label{sec:Results}
\input{06-Results}

\section{Conclusions}
\label{sec:conclusions}
\input{07-Conclusions}

\bibliographystyle{IEEEtran}


\end{document}

%% file: packages.tex
\usepackage[english]{babel}
\usepackage[utf8]{inputenc}

\usepackage{color,xcolor,ucs}

\usepackage{subfig}
\usepackage{floatrow}
\usepackage{tabularx}
\usepackage{float}
\usepackage{amsfonts}
\usepackage{helvet}         
\usepackage{courier}        
\usepackage{type1cm}        
\usepackage{amsmath}
\usepackage{amssymb}
\usepackage{makeidx}         
\usepackage{comment}         
\usepackage{graphicx}        
\usepackage{multicol}        
\usepackage[bottom]{footmisc}
\usepackage{bm}

\usepackage{cite}
\usepackage{url}

\usepackage{dsfont}

\usepackage[unicode=true, bookmarks=true,colorlinks=true]{hyperref}
\usepackage{xr-hyper}

\usepackage{algpseudocode,algorithm,algorithmicx}

\usepackage{xspace}
\usepackage{rotating}

\usepackage{graphicx}
\usepackage{threeparttable}
\usepackage{multirow}
\usepackage[font=scriptsize,labelfont=bf]{caption}

\usepackage{amsmath,mathtools}

\usepackage{geometry}

%% file: commands.tex
\newcommand{\prob}[1]{\ensuremath{\mathbb{P}({#1})}}

\algrenewcommand\algorithmicrequire{\textbf{Input:}}
\algrenewcommand\algorithmicensure{\textbf{Input:}}
\algnewcommand{\LineComment}[1]{\State \(\triangleright\) #1}

\makeatletter
\DeclareRobustCommand{\atan}{%
	\operatorname{atan}%
	\@ifnextchar2{_}{}%
}

%% file: title.tex
\title{Probabilistic Qualitative Localization and Mapping}

\author{authors}
\author{Roee Mor and Vadim Indelman 
	\thanks{R. Mor is with the Department of Computer Science,  Technion - Israel Institute of Technology, Haifa 32000, Israel. V. Indelman is with the Department of Aerospace Engineering, Technion - Israel Institute of Technology, Haifa 32000, Israel.   {\tt roeeki.mor@gmail.com vadim.indelman@technion.ac.il}. This work was  partially supported by the Israel Ministry of Science \& Technology (MOST).}
}

\date{}

%% file: 00-Abstract.tex
\begin{abstract}
	

Simultaneous localization and mapping (SLAM) are essential in numerous robotics applications, such as autonomous navigation. Traditional SLAM approaches infer the metric state of the robot along with a metric map of the environment. 
While existing algorithms exhibit good results, they are still sensitive to measurement noise, sensor quality, and data association and are still computationally expensive. 
Alternatively, some navigation and mapping missions can be achieved using only qualitative geometric information, an approach known as qualitative spatial reasoning (QSR). We contribute a novel probabilistic qualitative localization and mapping approach in this work. We infer both the qualitative map and the qualitative state of the camera poses (localization). For the first time, we also incorporate qualitative probabilistic constraints between camera poses (motion model), improving computation time and performance. Furthermore, we take advantage of qualitative inference properties to achieve very fast approximated algorithms with good performance.  In addition, we show how to propagate probabilistic information between nodes in the qualitative map, which improves estimation performance and enables inference of unseen map nodes - an important building block for qualitative active planning. We also conduct a study that shows how well we can estimate unseen nodes. Our method particularly appeals to scenarios with few salient landmarks and low-quality sensors. We evaluate our approach in simulation and on a real-world dataset and show its superior performance and low complexity compared to the state-of-the-art. Our analysis also indicates good prospects for using qualitative navigation and planning in real-world scenarios.

\end{abstract}

%% file: 01-intro.tex
Robotic and autonomous navigation has much impact on state-of-the-art applications in various domains. Image-based navigation and simultaneous localization and mapping (SLAM) are vital in this field.  
The SLAM problem has been extensively investigated in the past three decades (see \cite{Cadena16tro} for a recent survey of state-of-the-art approaches and challenges). In particular, highly-efficient open source SLAM software packages \cite{Dellaert12gt, Kuemmerle11icra, Agarwal16go, SalasMoreno13cvpr} have been developed and are gradually incorporated into real-world applications.
Lately, the problem of planning under uncertainty and active SLAM has also received considerable research attention (see \cite{Placed22arxiv} for a recent survey). 

Some challenges, however, remain. Firstly, while passive SLAM often achieves online performance, real-time performance for low-cost platforms is more challenging. In active planning, complexity is still an obstacle. Secondly, state-of-the-art approaches are mainly based on linearization of the non-linear geometric problem to use fast solvers \cite{Kaess12ijrr, Kaess08tro, Kuemmerle11icra, Polok13rss}. These approaches usually use many landmarks to enable noise filtering and outlier removal algorithms, another factor in high complexity and error accumulation. In many cases, these approaches require an accurate initial guess for the estimated variables, achieved using good GPS or IMU sensors or via accurate image-based camera re-sectioning techniques. \cite{Scaramuzza11ram, Fraundorfer12ram} is an overview of basic methods. Some advanced robust graph optimization techniques that attempt to be resilient, or less sensitive, to outliers are \cite{Carlone14iros, Lee13iros, Indelman16csm, Sunderhauf13icra, Olson13ijrr, Pathak18ijrr}. Non-parametric approaches to the SLAM problem (e.g.~\cite{Fourie16iros, Huang21icra}) try to overcome some of these issues by directly solving the nonlinear problem. These approaches, however, still need to be performed online.

A different approach is topological mapping, which considers relative attributes between different places. In this approach, a graph represents the environment with vertices for different places and edges for relative attributes (e.g., reachability between places). This problem is known as "Visual Place Recognition," as the estimation usually handles a discrete set of places. These approaches usually contain minimal or no geometrical data and do not fully integrate geometric inference. The advantage is that no continuous geometric estimation means no error accumulation and low noise dependency. On the other hand, the lack of geometric constraints in the estimation process usually makes the approaches computationally expensive. \cite{Lowry16arxiv} surveys topological mapping and place recognition approaches. 

Qualitative spatial reasoning (QSR) is yet another approach in which an agent perceives the environment through the geometric relation between objects. However, qualitative geometric relations are considered instead of metric locations and orientations. In addition, in many cases, instead of using a global frame of reference, local object-relative reference frames are used to relate to close-by landmarks. As discussed in this work, QSR is the base for our approach. While recent QSR works include \cite{Moratz2012AI},\cite{Mcclelland14jais} and \cite{Padgett17ras}, section \ref{sec:RelatedWork} provides a more detailed overview of QSR.


\subsection{Motivation}
\label{sec:Motivation}
\begin{figure}[]
	\centering
	\subfloat[]{\includegraphics[width=0.5\textwidth]{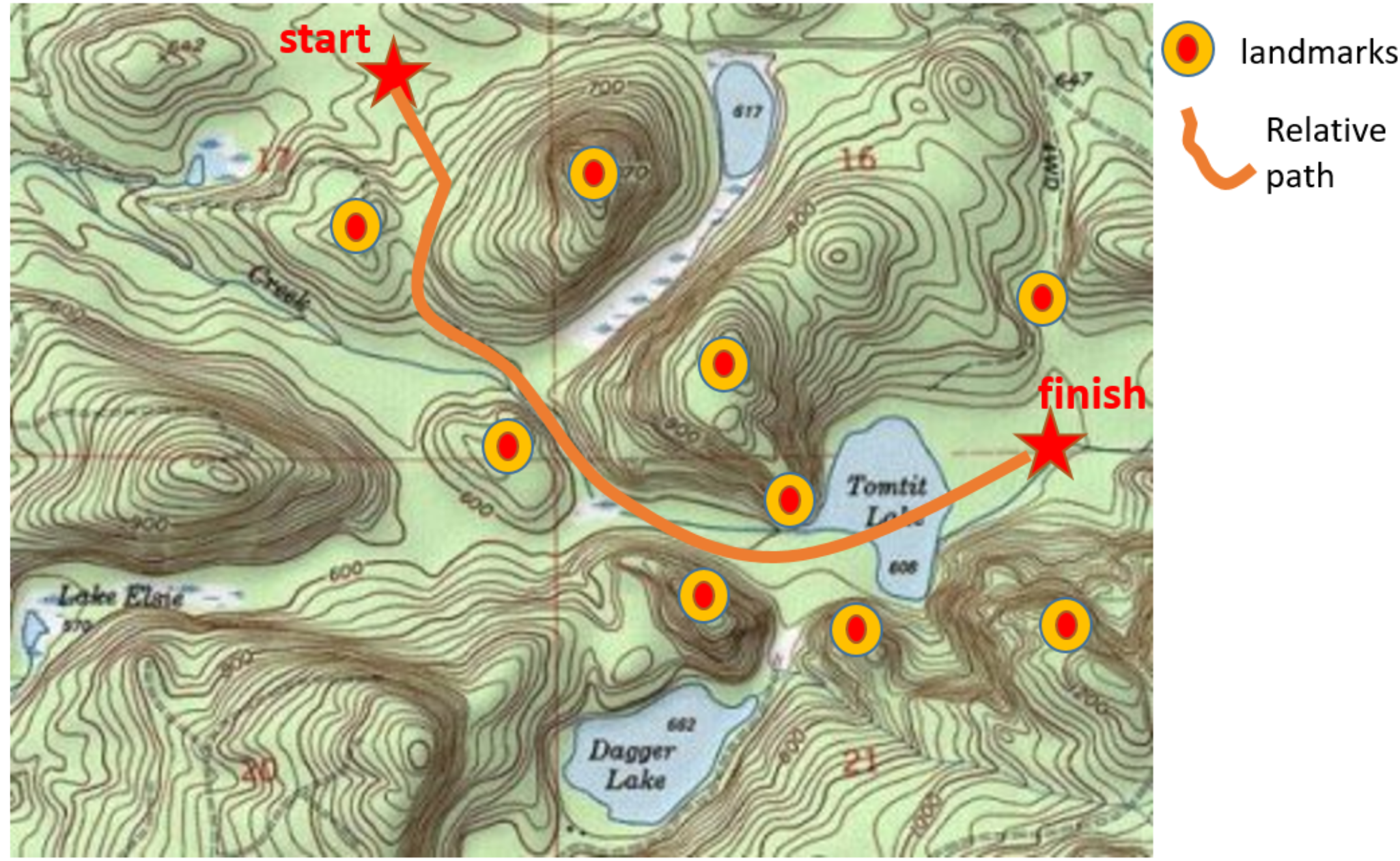}\label{fig:motivation1} }
	\subfloat[]{\includegraphics[width=0.5\textwidth]{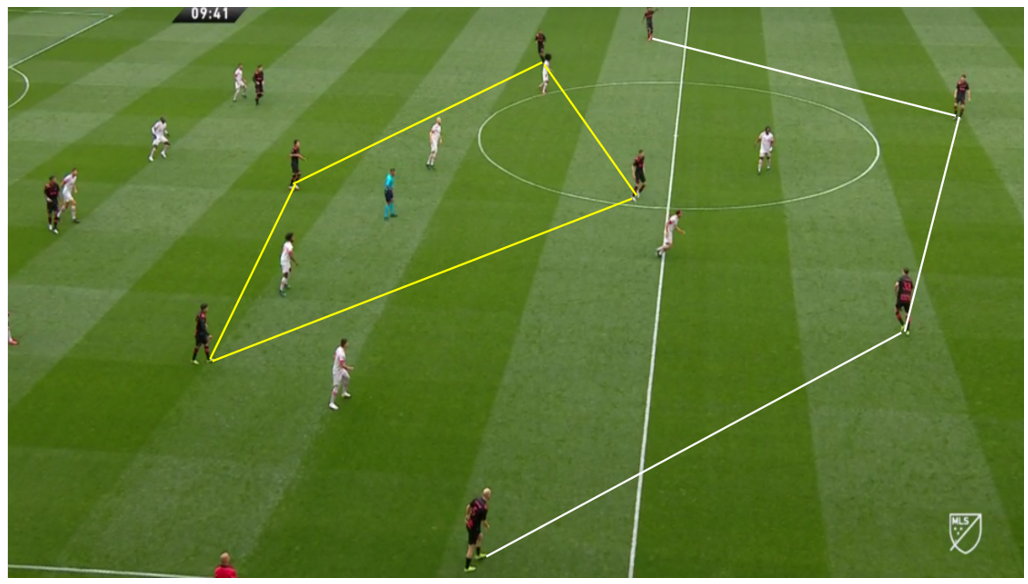}\label{fig:motivation2} }
	\caption{\label{fig:motivation}Qualitative geometry used by humans for various tasks: (a) planning route and navigating relative to landmarks and qualitatively using topographic map; (b) Football players plan and execute qualitative relative positioning. \vspace{-15pt}}
\end{figure}

QSR research was initially motivated by how humans and animals efficiently perform complex path planning and navigation tasks. They often relate to close landmarks instead of a large-scale global frame of reference and use a qualitative perception of the environment instead of an accurate metric perception (e.g., in figure \ref{fig:motivation}). In QSR approaches, estimation of the map and robot states is qualitative, hence, less metrically accurate but also less noise dependent. The general idea is that this more straightforward approach might be easier and more adequate for many robotic autonomous tasks. 

In traditional metric SLAM approaches, noise sensitivity demands using many landmarks for noise filtering, which partially accounts for the computational load. In addition, using many landmarks makes data association harder. 

Since qualitative inference is less sensitive to noise, it has the potential to be easier to implement. First, lesser noise sensitivity might require less noise filtering and, therefore, might be suited for using a smaller number of landmarks. Using a small number of salient landmarks can, in turn, improve landmark association. Secondly, since qualitative inference is less accurate, there is room for coarser approximations that produce simpler algorithms. These possibilities can lead to computationally light inference algorithms. In addition to the insight that many robotic tasks do not require accurate metric navigation, these properties motivate the research reported herein.

In this work, we contribute a probabilistic QSR-based mapping and localization framework designed for large-scale navigation with simple sensors and low complexity. Before stating the specific contributions of this work, we discuss the most relevant QSR approaches.

\subsection{Related Work}
\label{sec:RelatedWork}
\begin{figure}[]
	\centering
	\subfloat[]{\includegraphics[width=0.17\textwidth]{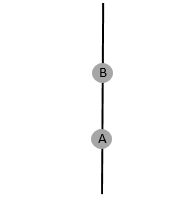}\label{fig:LR} }\quad
	\subfloat[]{\includegraphics[width=0.2\textwidth]{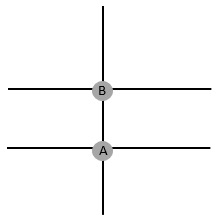}\label{fig:FDC} }\quad
	\subfloat[]{\includegraphics[width=0.2\textwidth]{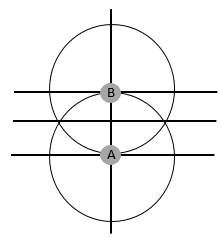}\label{fig:EDC} }\quad				
	\subfloat[]{\includegraphics[width=0.2\textwidth]{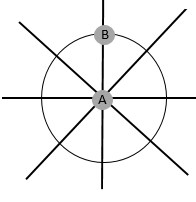}\label{fig:TPCC} }
	\caption{\label{fig:QSRmetrics}Different partitions of the metric space: (a) left right \cite{Scivos04icsc}; (b) Freksa's double cross \cite{Freksa92smc}; (c) Extended double cross \cite{Mcclelland14jais};  (d) TPCC \cite{Moratz2008Jovl}. \vspace{-15pt}}
\end{figure}

The application of QSR to robotic navigation and mapping started in the 90s. The early work by \cite{levitt1990AI} suggests qualitative localization of a robot about landmarks, given their azimuth ordering as seen by the agent in a single view. This approach has been extended in \cite{zheng1992ijcv,schlieder1993qrdt} and \cite{wagner2004ras} to include multi-view inference and some aspects of data association and place re-identification, but not a complete SLAM problem. Other methods address the qualitative representation of relative orientation between two oriented landmarks, such as "bi-pole orientation"\cite{Moratz2011AI} and "OPRAM"\cite{Moratz2012AI}. Another approach for representing spatial location was suggested by Freksa  \cite{freksa1992strg}. This approach localizes in a relative frame regarding two landmarks. Instead of considering the metric location in this frame, Freksa partitions the space into a discrete set of qualitative states known as "Freksa's single cross." The "Freksa's double cross" was suggested in \cite{Freksa92smc} and was recently extended by McClelland et al.~\cite{Mcclelland14jais} to the "extended double cross" (EDC) for a more detailed representation. Other qualitative partitions were also proposed over the years, including a "close and far pie" (TPCC) by \cite{Moratz2008Jovl} and "Left Right" by \cite{Scivos04icsc}.  Figure \ref{fig:QSRmetrics} illustrates these partitions.

The first comprehensive QSR-based mapping and navigation framework was proposed by McClelland et al.~\cite{Mcclelland14jais} for NASA's planetary rover. The authors proposed a qualitative algorithm for 2D large-scale mapping with a low-quality monocular camera and no GPS or ego-motion sensing. They use azimuth measurements and range ordering of each triplet of landmarks observed together to estimate one landmark location in a local frame defined by the other two. Instead of considering the metric location of the landmark, they use a discrete set of EDC-partitioned qualitative states. Geometric estimation classifies each qualitative state as "feasible" or "non-feasible" in a binary manner. The mapping goal is to infer all feasible qualitative states for each landmark triplet. Measurements from different viewpoints can reduce ambiguity. An additional "composition" stage propagates qualitative data through triplets with common landmarks, including triplets that were never viewed together by the agent.

Further work \cite{Padgett17ras} extends this method to probabilistic estimation. Instead of assigning true or false labels, they infer the probability for each qualitative state of each landmark triplet. \cite{Padgett18ras} takes another step and addresses active qualitative planning. 

However, \cite{Mcclelland14jais} and \cite{Padgett17ras} do not model any spatial connection between different camera views of the same triplet (such as motion model or geometric triangulation). They also do not address probabilistic "composition." Finally, in both papers, the focus is on mapping, not camera localization.
The work of \cite{zilberman2022incorporating} on the other end specifically addresses the use of qualitative composition between landmark triplets in QSR problems. It focuses on the optimal sequence of compositions for propagating data in a qualitative map given a specific source and target nodes. It also addresses the question of which new nodes can be added to the map by composition.

\subsection{Contributions}

In this work, we take a few steps closer to a full probabilistic qualitative framework for localization and mapping. We aim for simple sensors (such as a low-quality monocular camera with no GPS or significant IMU), low complexity, and large-scale navigation. The potential of qualitative geometry for simple and light inference (as mentioned in \ref{sec:Motivation}), and the insight that many robotic tasks do not require accurate metric navigation, drive us to extend existing QSR approaches. Moreover, we envision our approach as a step towards active QSR planning, leveraging belief space planning formulation (e.g.~\cite{Indelman15ijrr, Kopitkov17ijrr, Pathak18ijrr}). We use a different and innovative formulation of the problem to address some of the key limitations and introduce several key improvements to state-of-the-art, most notably \cite{Mcclelland14jais} and \cite{Padgett17ras}.  

In particular, our \emph{main contributions} are as follows:

\begin{enumerate}
	\item While state-of-the-art QSR approaches focus mainly on mapping, we develop a holistic probabilistic QSR approach that also addresses localization.
	\item We incorporate a motion model to improve both performance and complexity.
	\item We develop a global non-linear solver with a simple algorithmic flow. It is better suited for various measurement types and does not require prior knowledge. We also utilize qualitative inference robustness to develop a very fast approximated algorithm with similar performance.
	\item We develop probabilistic composition - a novel way for efficiently propagating information in the qualitative map, extending the deterministic approach in \cite{Mcclelland14jais}.  Probabilistic composition is key for active qualitative planning, as shown in our follow-up work \cite{Zilberman22iros}.
	\item We formulate a discrete factor graph representation for the qualitative map and use it for propagating information. The full solution for getting marginals by elimination is computationally infeasible. Therefore, we develop an approximate fast and effective algorithm.
	\item  We analyze how propagating qualitative information in the factor graph is related to graph topology and prior information. For this purpose, we suggest new information-related metrics and a simplified model that gives an empirically strong correlation to actual qualitative data propagation.
	\item  We evaluate the performance of our approach in simulation and on a real-world dataset and compare it against the state-of-the-art.
	\item  We also made available an open-source code repository \cite{qsrslam} that implements all parts of our work. This repository holds a python-based implementation meant for performance analysis and to be a reference for future work. It is only partially real-time optimized.
\end{enumerate}

The present paper is an extension of the work presented in \cite{Mor20iros}, where we introduced our  innovative approach for the qualitative localization and mapping problem.  The formulation presented in the current manuscript is more general and includes more extensive evaluations. As further core contributions, we
address the map as a qualitative factor graph and provide insights about how much data can be propagated (items 5 and 6 above). 	In this paper, we bring a full and comprehensive formulation and analysis of our approach.


This paper's organization is as follows. Section \ref{sec:problem-formulation} introduces notations and provides problem formulation. Section \ref{sec:Approach} describes in detail our approach. It presents a probabilistic formulation of the qualitative localization and mapping problem. Then it presents our proposed algorithm for the 2D case in detail while also addressing run-time aspects. Lastly, it also includes the derivation of our probabilistic composition technique and how to use it in a factor graph for propagating information in a qualitative map. Section \ref{sec:Results} provides performance evaluation. It also includes an analysis of the relations between propagating qualitative information in the map to its factor graph topology. Section \ref{sec:conclusions} concludes the discussion and suggests several avenues for future research.

%% file: 02-ProblemFormulation.tex
We consider a robot navigating in and mapping an unknown environment. As the robot moves, it tracks landmarks across different image frames.  We aim to \emph{qualitatively} describe the environment and camera trajectory. Motivated by the approach in \cite{Mcclelland14jais},\cite{Padgett17ras}, we consider multiple relative landmark-centric frames of reference and use qualitative geometry to describe camera and landmark locations in these frames.

\begin{figure}[]
	\centering
	\subfloat[]{\includegraphics[width=0.30\textwidth]{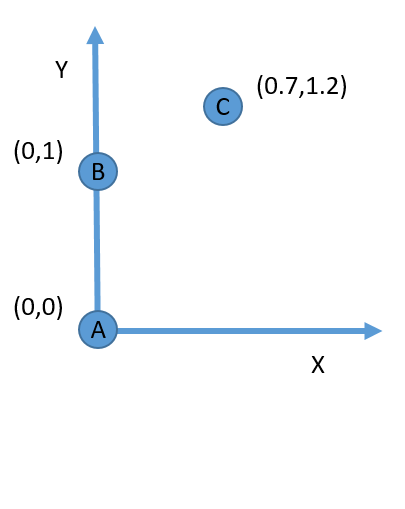}\label{fig:ABframe1} }\quad
	\subfloat[]{\includegraphics[width=0.30\textwidth]{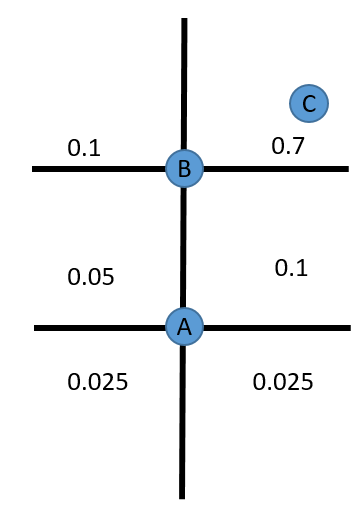}\label{fig:ABframe2} }\\	
	\caption{\label{fig:ABframe} (a) AB landmark relative metric frame of coordinates. (b) AB frame qualitative state probability distribution $\prob{S^{AB:C}}$ (Freksa's double cross in this example).}
\end{figure}

While our formulation is general, we test our approach on the 2D case. For every three landmarks $A,B,C$  observed together, we set a landmark-centric local frame, so that landmark A location is $L^A=(0,0)$ and landmark B location is $L^B=(0,1)$  (see Figure \ref{fig:ABframe1}). We describe the qualitative location of landmark $C$ in this frame using a discrete space partition into a set of $m$ qualitative states (see Figure \ref{fig:ABframe2}). The qualitative state of landmark $C$ is $S^{AB:C}$. It is a vector of dimension $m\times1$  that contains the hypothesis for $C$ being in each qualitative state. We denote the event that $C$ is in a specific a qualitative state $S^{AB:C}= i;\in\{1,...,m\}$ as $s^{AB:C}_i$. We also denote the metric location of landmark $C$ as $L^{AB: C}$. 

Similarly, the metric location of the camera in the $AB$ frame at time step $n$ is denoted by $X^{AB}_n$, and the corresponding camera qualitative state is denoted by $S^{AB:X}_n$.  The camera qualitative location in the entire map at time step $n$ is then the collection of its states regarding all local frames:  $S^X_n=\{S^{i,j:X}_n\}$. 

One can choose the space partition to fit different tasks, platforms, or scenarios (see Figure \ref{fig:QSRmetrics}). In our implementation and tests, we employ the EDC partition used in previous works \cite{Padgett17ras}, \cite{Mcclelland14jais}.

The true mapping and localization qualitative states ($S^{ij:k}$ and $S^{i,j:X}_{1:n}$) are unknown. In this work, we infer them within a Bayesian framework using a set of measurements $Z^{ABC}_{1:n} =  \{ Z^{ABC}_j; j \in \{1,...,n\} \}$ from time steps $1,...,n$. The  observation of landmarks $A,B,C$ from time instant $j$ is $Z^{ABC}_j$. We note that while, for simplicity, this notation suggests observations are assumed to exist for all time instances $[1,n]$, in practice, our method does not require this assumption, as further elaborated in Section \ref{sec:Composition}.

Herein, we also introduce an innovative usage of a motion model within a qualitative formulation, which extends previous work. As will be seen, this enables us to use more substantial geometric constraints to improve estimation.  Thus, we assume action $a_{n-1}$ for moving the camera between time instances $n-1$ and $n$.

Our goal at time instant $n$ is to estimate the posterior probabilities of qualitative landmark C and camera states, given history $H^{ABC}_n \doteq \{ Z^{ABC}_{1:n},a_{1:n-1} \}$:
\begin{equation}
	\prob{S^{AB:C}| H^{ABC}_n} \ \ , \ \ \prob{ S^{AB:X}_{1:n}| H^{ABC}_n }.
	\label{eq:QSprobs}
\end{equation}

The formulation in this paper is general and can use any measurement and motion model (although we use the Markov assumption). We consider 2D coordinate systems and a monocular camera setup in our implementation and results. Therefore landmark 2D location is $L^{AB:C}=(x^{AB:C},y^{AB:C})$ and camera pose 2D is $X^{AB}_n=(x^{AB:X}_n,y^{AB:X}_n,\alpha^{AB:X}_n)$, where $\alpha$ is the camera orientation angle. Measurements are bearing angles to landmarks $A,B$ and $C$, i.e.~$Z^{ABC}_n=\{\phi^A_n,\phi^B_n,\phi^C_n\}$.

For this setup, we assume a Gaussian measurement model for $i\in\{A,B,C\}$:
\begin{equation}\label{eq:MsrLkl}
	\prob{\phi^i_n | L^i, X_n}\propto \exp\left\{-\frac{1}{2}\Vert \phi_n^i - f(L^i,X_n) \Vert^2_{\Sigma_v} \right\},
\end{equation}	
where $\Sigma_v$ is the measurement noise covariance and 
\begin{equation} \label{eq:f}
	f(L^i,X_n) \doteq  \arctan(\frac{y^i-y^X_n}{x^i-x^X_n}).
\end{equation}
Considering bearing measurements to different landmarks statistically independent, the joint likelihood for $Z^{ABC}_n$ is readily obtained as a product of individual likelihood terms (\ref{eq:MsrLkl}) for each bearing measurement $\phi_n^i\in Z^{ABC}_n$.

In our setup, we consider a simple motion model that can be used with a monocular camera and basic azimuth-keeping control. Specifically, we consider the robot is moving along a specific heading $a_n=\psi_n$ relative to the previous camera pose and assume Gaussian noise as in
\begin{equation}\label{eq:MotionModel}
	\prob{\!X^{AB}_n | X^{AB}_{n-1}, a_{n-1}\!} \! \propto 	\! \exp \! \left\{\!\! -\frac{1}{2}\Vert a_{n-1} \! - \!  g(X^{AB}_n \!, \! X^{AB}_{n-1}) \Vert^2_{\Sigma_w} \! \right\}\!\!,
\end{equation}
where $\Sigma_w$ is the motion (process) noise covariance and $g(.)$ is defined, similarly to Eq.~(\ref{eq:f}), as  $g(X_n , X_{n-1}) \doteq  \arctan(\frac{y^X_n-y^X_{n-1}}{x^X_n-x^X_{n-1}})$. The motion model (\ref{eq:MotionModel}) does not constrain the camera orientation. As will be seen, incorporating a motion model, even as simple as this, leads to a much better qualitative state estimation.

The second part of our work (section \ref{sec:ApproachComposition}) addresses the entire qualitative map of all landmarks and how to propagate information between landmark triplets in this map. The environment is a set of $m$ landmarks; we denote the group of all possible landmark triplets as $M_{all}=\{S^{i,j:k}\}_{i,j,k=1...m; j\neq i, k\neq j,i}$. The qualitative map is then a subset of the triplets that we are interested in $M=\{S^{i,j:k}\} \subseteq M_{all}$. These can be triplets that have been observed or otherwise ones we want to estimate (see Figures \ref{fig:map1} and \ref{fig:map2}).

In section \ref{sec:ApproachComposition}, we also approach the propagation of information in the map. Propagating information is the process of inferring a specific triplet state given its history and the history of other triplets: $\prob{S^{AB:D}|H^{AB:C},H^{BC:D},H^{AB:D}}$, or alternatively given only the estimation of other triplets qualitative state $\prob{S^{AB:D}|S^{AB:C},S^{BC:D},S^{AB:D}}$. In this section, we will use a shortened notation to refer to triplets to simplify formulation. Instead of specifying the two frame landmarks and the third one: AB:C, we just refer to the triplet as $ti$ for $i=1,2,...$; so for example,  $AB:C\equiv t1, BC:D\equiv t2, CD:E\equiv t3$ and this way we annotate $S^{AB:C},S^{BC:D},S^{CD:E}$ as $S^{t1},S^{t2},S^{t3}$  and $H^{AB:C}, H^{BC:D}, H^{CD:E}$ as $H^{t1},H^{t2},H^{t3}$. In this notation a map with $k$ triplets is notated as $M=\{S^{ti}\} ;i=1...k$.

%% file: 03-Approach.tex
In this work, we present our probabilistic qualitative localization and mapping approach. The first part of our work (section \ref{sec:ApproachSingleTriplet}) discusses the single triplet problem. First, we generally formulate the probabilistic inference of camera and landmark triplet qualitative states when seen from multiple views and consider a motion model. We then address the 2D problem as a test case in more detail. We analyze the problem geometry and describe our unique qualitative inference algorithms (sections \ref{sec:AlgorithmDesignConsiderations} and  \ref{sec:-DetailedAlgorithm}).

The second part of our work (section \ref{sec:ApproachComposition}) addresses data propagation between different triplets in the qualitative map. We derive a novel probabilistic composition algorithm for propagating information between different landmark triplets in the qualitative map. We formulate it using a factor graph representation. Then, we analyze the dependency of propagated information on the graph topology and prior knowledge. We use unique information metrics and graph propagation algorithms.

Section \ref{sec:Results} includes test results and analysis for both parts of our approach and comparison to state-of-the-art.

%% file: 04-approach_single_triplet.tex
\vspace{5pt}

As specified in Section \ref{sec:problem-formulation}, our approach considers multiple landmark-centric triplet frames. In this section, we focus on a single triplet of landmarks $A,B$, and $C$ viewed together from multiple views (at time steps $1,\ldots,n$). In section \ref{sec:Probabilistic navigation and mapping}, we infer camera and landmark qualitative states given multiple views of the landmarks and a motion model. Then, we address the 2D SLAM problem as a test case in more detail. We discuss the geometric properties of the 2D problem (section \ref{sec:geometry}) and present our unique algorithm for solving it in detail. In addition, we develop a very fast approximated algorithm that achieves good performance utilizing the advantages of qualitative inference (sections \ref{sec:AlgorithmDesignConsiderations} and  \ref{sec:-DetailedAlgorithm}).

\subsection{Probabilistic Formulation}
\label{sec:Probabilistic navigation and mapping}
\input{04a-Formulation}

\subsection{Basic Problem Geometry}
\label{sec:geometry}
\input{04b-Geometry}

\subsection{Algorithm Design Considerations}
\label{sec:AlgorithmDesignConsiderations}
\input{04c-AlgorithmDesignConsiderations}

\subsection{Detailed Algorithm}
\label{sec:-DetailedAlgorithm}
\input{04d-DetailedAlgorithm}

\subsection{Faster Variant Algorithm}
\label{sec:faster variant}
\input{04e-FasterVariantAlgorithm}

%% file: 04a-Formulation.tex
Let us look at a single triplet of landmarks $A,B$, and $C$ viewed together at time steps $1,\ldots,n$. When considering a specific $AB:C$ local frame, we aim to estimate the qualitative state of landmark C and the camera at each time step using measurements (\ref{eq:QSprobs}). The state-of-the-art \cite{Padgett17ras} directly formulates this problem, considering only landmark to camera measurements. We notice that using other types of information may improve inference and therefore adopt a different approach.

The qualitative problem is derived from an underlying fundamental metric SLAM problem. This underlying problem is a small three landmark multiple view SLAM problem of determining camera poses and landmark C location, given a set of noisy measurements $\prob{X^{AB}_{1:n},L^{AB:C}|Z^{ABC}_{1:n}}$. It is well known that incorporating a motion model makes the problem easier to solve and requires fewer measurements and less prior knowledge. This insight is also valid for the qualitative problem.  We, therefore, take a more general formulation than the one in \cite{Padgett17ras}, which enables us to naturally translate the effect of the motion model into the qualitative problem. We do this by formulating the qualitative problem directly related to the underlying metric problem.

Given the landmark measurements \eqref{eq:MsrLkl} and motion model \eqref{eq:MotionModel} for each camera transition, we want to infer the posterior probabilities of the landmark $C$ qualitative state and camera qualitative trajectory (\ref{eq:QSprobs}), both in the $AB$ frame. To reduce clutter, we drop the superscript ${AB}$ notation in this section as long as everything is in the $AB$ frame. Also, for an easier explanation, instead of looking at $\prob{ S^C | H_n}$ we look at the separate components of this random vector: $\prob{ s_i^C | H_n}$ (i.e., the probabilities of the landmark  $C$ to be in each qualitative state separately). Generalizing to $\prob{ S^C | H_n}$ is trivial.

We formulate $\prob{ s_i^C | H_n}$ through the underlying metric problem. To do so, we start by marginalizing over the metric camera poses and landmark locations, writing the belief over $s^C_i$ as:
\begin{equation}
	\prob{ s_i^C | H_n} = \iint\displaylimits_{X_{1:n},L^C}{  \prob{ s_i^C,X_{1:n},L^C| H_n }  dL^C dX_{1:n}}.
\end{equation}
We shall now apply chain rule:
\begin{equation}
	\!\begin{multlined}[t]
		\prob{ s_i^C | H_n} \!\!=\!\!\!\!\!\! \iint\displaylimits_{X_{1:n},L^C}{  \!\! \!\!\prob{ s_i^C|X_{1:n},L^C\!\!, H_n } \prob{ X_{1:n},L^C| H_n }} 
			dL^C dX_{1:n}.
	\end{multlined}
\end{equation}	
Note that landmark C metric location $L^C$ uniquely determines its qualitative state $s_i^C$ so that $\prob{s_i^C|L^C}=1$ for $L^C \in s_i^C$ and 0 else. Since $\prob{s_i^C|L^C}$ is independent of any other history and can also be replaced by the corresponding integration range, we get:
\begin{align}
	\prob{s_i^C | H_n} &= \iint\displaylimits_{X_{1:n},L^C}{  \prob{ s_i^C|L^C} \prob{ X_{1:n},L^C| H_n } dL^C dX_{1:n}} \nonumber \\
	&= \iint\displaylimits_{L^C \in s_i^C, X_{1:n}}{ \prob{X_{1:n},L^C| H_n}  dL^CdX_{1:n}}.
	\label{eq:slam1}
\end{align}
%
Similarly, inferring the camera's qualitative state:  
\begin{equation}\label{eq:slam2}
	\prob{ s_i^{X_i} | H_n} =  \!\! \iint\displaylimits_{X_i \in s_i, X_{1:n /i},L^C}{\!\!\!\! \prob{X_{1:n},L^C| H_n}  dL^CdX_{1:n}}.
\end{equation}

We get an intuitive result: solve $\prob{X_{1:n},L^C|H_n}$ - the corresponding SLAM problem  in the $AB$ frame and marginalize over camera trajectory $X_{1:n}$ and landmark C metric locations $L^C$ that belong to the relevant qualitative state. 

This approach is very different from previous works \cite{Padgett17ras}, \cite{Mcclelland14jais}. It has several advantages: (i) Summing over qualitative states can be trivially adjusted to any space partition (see Figure \ref{fig:QSRmetrics}). (ii) Solving the \emph{small} metric SLAM problem can be done using any existing method or code to fit different applications or scenarios.

We now further break down the SLAM problem $\prob{X_{1:n},L^C|H_n}$  into simpler factors using a standard SLAM formulation and show how to in-cooperate a motion model. Applying  Bayes' theorem and recalling the measurement 
model $\prob{Z_n|X_n,L_C}$  is independent of history $H_n$ gives:
\begin{equation}\label{eq:slamst1}
	\prob{X_{1:n},L_C|H_n} \! = \! \frac{1}{\zeta_n}  \prob{Z_n|X_n,L_C}\\
	\prob{X_{1:n},L_C|H_{n}^-}.
\end{equation}	
Where $H_n^-\doteq \{a_{1:n-1},Z_{1:n-1}\}$ is the history without measurements from current time, such that $H_n=H_n^- \cup \{Z_n\}$. In addition, $\zeta_n \doteq \prob{Z_n|H_n^-}$ is independent of integration variables, so we can normalize outside the integrals.

We proceed by using the formula of total probability over $X_n$, and the Markov property of the motion model to get the following recursive formulation:
\begin{equation}\label{eq:slamst2}
	\prob{X_{1:n},L_C,H_n} = \frac{1}{\zeta_n}  \prob{Z_n|X_n,L_C}   \prob{X_n|X_{n-1},a_{n-1}} \nonumber \prob{X_{1:n-1},L_C|H_{n-1}}.
\end{equation}
%
Repeating these two steps $n-1$ times we get
\begin{equation}\label{eq:slamst3}
		\prob{X_{1:n},L_C|H_n} = \frac{\prob{Z_1|X_1,L_C} \prob{X_1,L_C}}{\prob{Z_1}}
		\prod_{i=2}^n{\frac{1}{\zeta_i}  \prob{Z_i|X_i,L_C} } \prob{X_i|X_{i-1}, a_{i-1}},
\end{equation}	
where $\zeta_i \doteq \prob{Z_i|a_{1:i-1},Z_{1:i-1}}$ is the normalization factor. 
Using this standard decomposition, we can see how measurement and motion models can solve the underlying SLAM problem.

Recalling Eqs.~\eqref{eq:slam1}  and \eqref{eq:slam2}, we can now integrate over the camera trajectory and landmark C location to estimate the required qualitative states.

%% file: 04b-Geometry.tex
After formulating the problem, we look at the 2D SLAM problem as a test case. In the following sections (\ref{sec:AlgorithmDesignConsiderations}, \ref{sec:-DetailedAlgorithm}), we suggest novel algorithms for solving this problem while taking advantage of qualitative geometry's unique properties to simplify the inference. This section lays the basics for understanding our motivation and choices. We analyze the geometric properties of the 2D problem and the effect of incorporating a motion model on the solution quality.

As specified in Section \ref{sec:problem-formulation}, our approach considers multiple landmark-centric triplet frames. For each local frame, the underlying fundamental metric problem $\prob{X^{AB}_{1:n},L^{AB:C}|Z^{ABC}_{1:n}}$ determines camera poses and landmark C location in the local AB frame, given a set of measurements. When considering measurement noise, the problem is a small metric three landmark - multi-view SLAM problem. This SLAM problem is observable only with enough landmark measurements and camera poses or with priors. The thing is that achieving enough measurements or priors for each local frame can be problematic in some robotic platform scenarios.

While the state-of-the-art only uses measurements, we consider the extended problem $\prob{X^{AB}_{1:n},L^{AB:C}|Z^{ABC}_{1:n},a_{1:n-1}}$ and incorporate the corresponding motion model. The motion model enables solving the problem better and with fewer measurements. A simple degree of freedom analysis can demonstrate this for the 2D  case. The unknowns are landmark $C$ location $(x_C,y_C)$, and camera poses $(x_{1:n},y_{1:n},\alpha_{1:n})$. For $n$ time steps, we have $3n+2$ unknowns. When using only azimuth measurements $ (\phi^A_{1:n},\phi^B_{1:n},\phi^C_{1:n})$ to landmarks $A,B$ and $C$, the number of equations is $3n$. This problem is, therefore, under-determined. When considering actions $a_{1:n-1}$ (i.e.~the azimuth angles from one camera pose to the next, $\psi_{1:n-1}$) and the corresponding motion model \eqref{eq:MotionModel}, 
the number of equations is $4n-1$. For $n \geq 3$, the number of equations equals or exceeds the number of unknowns, and the problem becomes fully observable. The intuition is that incorporating a motion model enables solving landmark $C$ location by triangulating line-of-sight vectors from multiple views.


This well-known insight related to the metric problem is also valid for the qualitative problem. Therefore, incorporating a motion model will solve the qualitative problem with fewer measurements and prior knowledge. In sections \ref{sec:-DetailedAlgorithm} and \ref{sec:SingleTripletResults} we will look at the effect in more detail.

We now dive into the 2D problem basic geometry to better understand our algorithm and the effect of using a motion model in sections \ref{sec:-DetailedAlgorithm} and \ref{sec:faster variant}. We look at the basic deterministic geometry, so in this explanation, we consider the ideal noise-free measurement and motion models that we shall denote by $\bar{Z}$.

\begin{figure}[]
	\centering
	\subfloat[]{\includegraphics[width=0.3\textwidth]{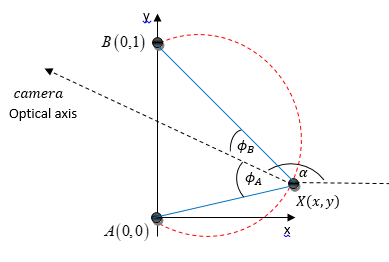}\label{fig:2pResecsubfig1} }\quad
	\subfloat[]{\includegraphics[width=0.3\textwidth]{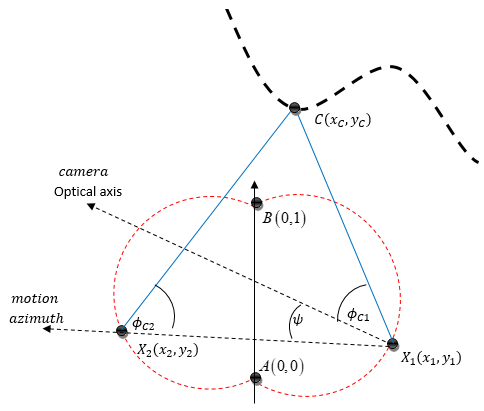}\label{fig:2pResecsubfig2} }\quad
	\subfloat[]{\includegraphics[width=0.3\textwidth]{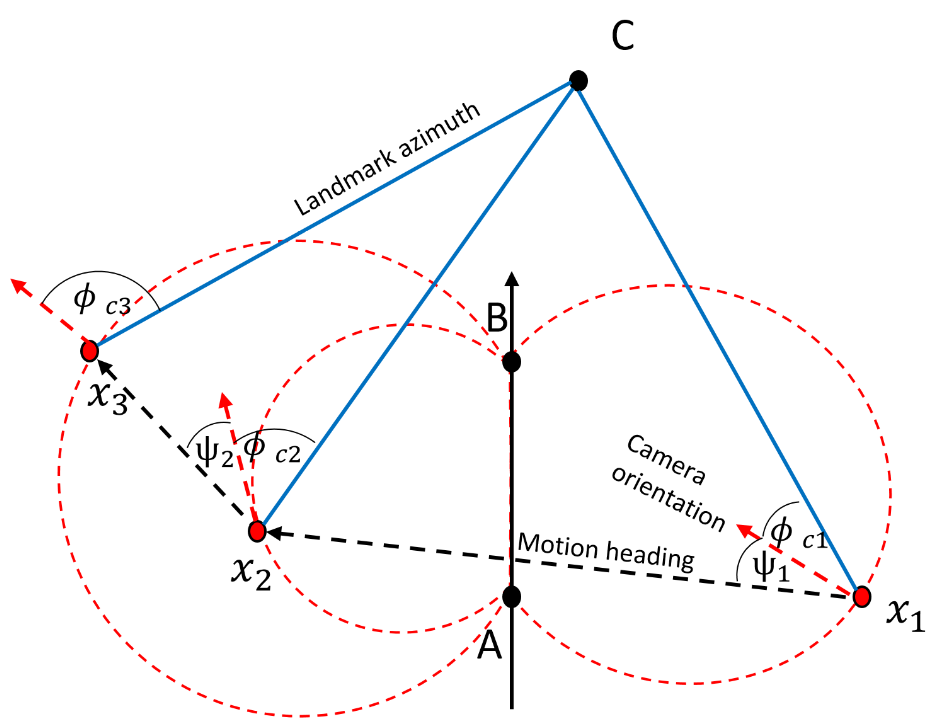}\label{fig:2pResecsubfig3} }\\	
	\caption{\label{fig:2pointresection} (a) 2D 2-point camera resection from single view. (b) camera resection and landmark triangulation with two views. (c) camera resection and landmark triangulation with three views.}
\end{figure}

\begin{figure*}[]
	\centering
	\subfloat[]{\includegraphics[width=0.3\textwidth]{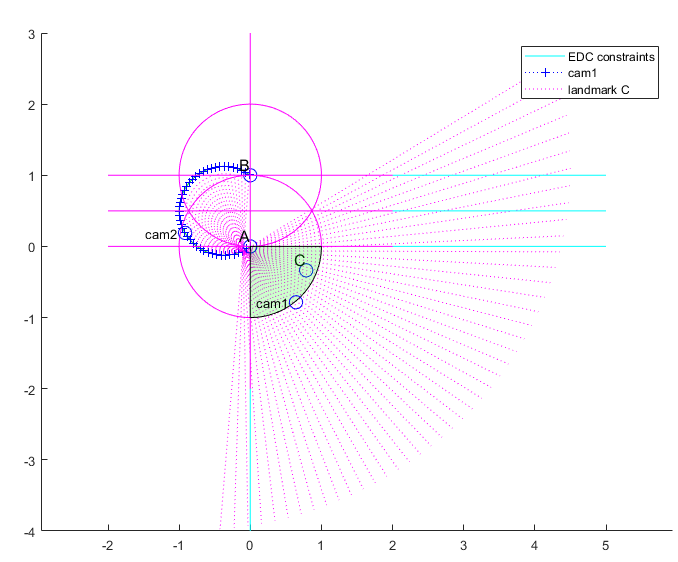}\label{fig:DeterministicEx1} }\quad
	\subfloat[]{\includegraphics[width=0.3\textwidth]{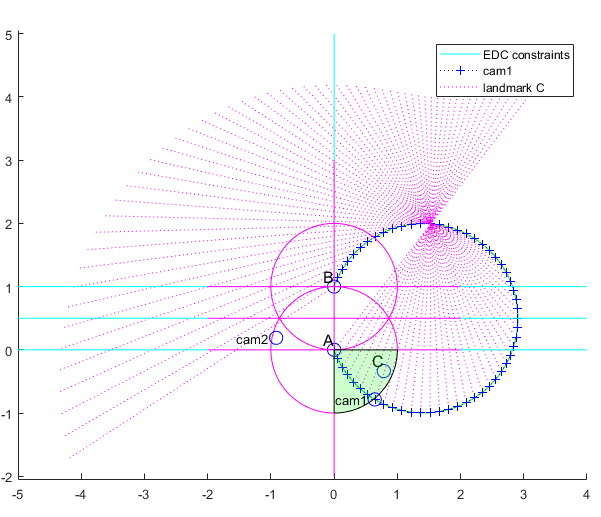}\label{fig:DeterministicEx2} }\quad			\subfloat[]{\includegraphics[width=0.3\textwidth]{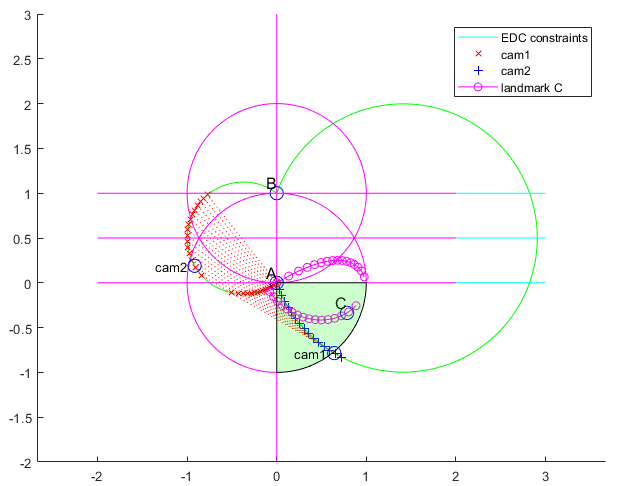}\label{fig:DeterministicEx3} }\\	
	\caption{\label{fig:DeterministicExmpl} Simulation example for first camera resection and landmark estimation: Actual camera positions and landmark positions are marked in circles. (a) First view. Estimated camera position marked by blue +. Estimated landmark C location marked in magenta (.) - see the legend. (b) Second view. Same markings. (c) Estimation by both views using motion model. Estimated camera position marked by +,x. Estimated landmark C location marked in magenta circles (see legend). }
\end{figure*}

First, we regard the single-view case. Landmarks A and B are known (we work in the AB frame), while the camera pose and landmark C are unknown. Determining camera pose by 2D azimuth measurements to two known landmarks is a well-known 2D 2-point camera resection problem. \cite{pierlot2014tro} specifies a complete and efficient analytic solution to this problem. Given these azimuth measurements, the camera's location must be on a \emph{circle} that goes through the two landmarks A and B. \cite{pierlot2014tro} gives the equations for the "locus circle" center and radius. Azimuth ordering of $\phi^A_n,\phi^B_n$ further confines the camera location to the left or right to the AB vector, see Figure \ref{fig:2pResecsubfig1}. It is also easy to show that assuming a specific camera location on the locus circle directly determines the camera orientation $\alpha_n$: 
\begin{equation} \label{eq:orientation}
	\alpha_n=\arctan(\frac{y_n^X}{x_n^X})+\phi_A=\arctan(\frac{y^X_n-1}{x^X_n})+\phi_B.
\end{equation}
Per every possible camera pose, landmark C can be located somewhere on the line of sight corresponding to the azimuth measurement $\phi^C_n$. A diagram illustrating these aspects is given in Figure \ref{fig:2pResecsubfig1}, while Figures \ref{fig:DeterministicEx1} and \ref{fig:DeterministicEx2} have a stimulative example. A crucial insight is that even in this deterministic setting, the problem is not fully observable, i.e.~solutions for metric camera poses and landmark locations have a continuous distribution. We denote this joint distribution by
$\prob{X^{AB}_n,L^{AB:C}|\bar{Z}^{ABC}_n}$.

If measurements from several time instances are available, the joint pdf is:
\begin{equation}
	\prob{X^{AB},L^{AB:C}|\bar{Z}^{ABC}_{1:n}}= \! \prod^n_{i=1}{\prob{X^{AB},L^{AB:C}|\bar{Z}^{ABC}_i}}.
	\label{eq:pdf_determ}
\end{equation}
The posterior distribution over qualitative states can be extracted by integration.

Now we examine the impact of incorporating a motion model. The second camera location $(x_2^*,y_2^*)$ is determined by the combination of the first camera pose $(x_1^*,y_1^*,\alpha^*_1)$, the action $\psi_1$, and the second measurement $Z_2=\{\phi^A_2,\phi^B_2,\phi^C_2\}$ (obtained after executing the action). It is the intersection of the line of motion with the valid part of the second measurement locus circle (see Figure \ref{fig:2pResecsubfig2}). The intersection can occur once, twice, or not at all. As a result, some of the first camera poses are disqualified. Another consequence is geometric ambiguity: some measurements can support two solutions for the second camera pose (and the location of landmark C correspondingly).

Given the two camera poses $(x_1^*,y_1^*,\alpha_1^*),(x_2^*,y_2^*,\alpha_2^*)$ and azimuth measurements $\phi^C_1,\phi^C_2$ to landmark C, its location $(x_c^*,y_c^*)$ can be triangulated. Considering all possible poses for the first camera reduces the possible locations landmark C to a curve (might be split into two curves in case of geometric ambiguity) - see Figure \ref{fig:2pResecsubfig2}. A simulation example that displays how our motion model allows disqualifying a part of the first camera poses and triangulating the landmark C - see Figure \ref{fig:2pResecsubfig2}.

When considering three or more measurements, only one or a few discrete possible locations for landmark C and the corresponding camera trajectories are left. Triangulation consistency further disqualifies most camera poses and landmark locations. (Figure \ref{fig:2pResecsubfig3}). The surviving landmark C locations are an intersection of the curve estimates for pairs of consecutive views. 

Using the motion model significantly increased the quality of the solution for the metric problem. In the probabilistic case, the probability for qualitative states should also improve as it is an integration on the metric pdf. This fact demonstrates the motivation for our work. 

We also notice that the geometric ambiguity enables several distinct solutions to the problem. Therefore probabilistic solutions based on linearization might converge to local minima and achieve subpar results. A global solution or hypothesis-based approach is therefore needed.

%% file: 04c-AlgorithmDesignConsiderations.tex
Standard SLAM approaches usually solve big bundle adjustment problems with many landmarks and camera poses. Achieving a global solution for such problems is hard. Therefore, It is common to model the estimated variables as multi-variate Gaussian and resort to linearization-based local optimum. These approaches require enough measurements and prior knowledge or an initial guess for the solution.

\begin{figure}[]
	\centering
	\subfloat[]{\includegraphics[width=0.35\textwidth]{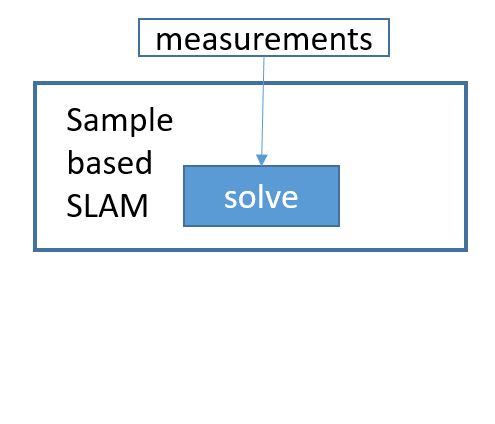} }\quad
	\subfloat[]{\includegraphics[width=0.35\textwidth]{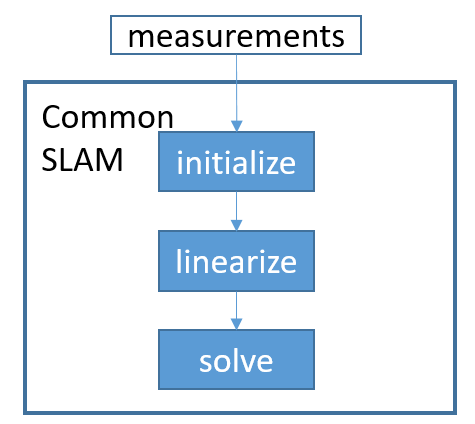} }\\
	\caption{\label{fig:algorithmic_flow} (a) general algorithmic flow of standard multi-variate Gaussian linearization based algorithm. (b)  general algorithmic flow of direct global sample based  algorithm.}
\end{figure}

Alternatively, we solve many small landmark-centric SLAM problems with three landmarks, and a few camera poses at a time. We also want to work with a small number of measurements and no prior knowledge. Under these conditions, we choose a sampling-based approach to solve the \emph{global} non-linear \emph{small} SLAM problem. This approach avoids linearization and gets a robust global solution. In addition, it does not need prior information or an initialization stage. Also, integration over coarse resolution qualitative states compensates for some sampling error. This robust and much simpler algorithmic flow is well suited for low-compute platforms (see figure \ref{fig:algorithmic_flow}).

This approach can be computationally expensive for a large SLAM problem, but it is feasible for many small problems. Furthermore, as discussed in chapter \ref{sec:geometry}, using a motion model and triangulation allows us to disqualify inconsistent samples rapidly, drastically reducing valid samples for two or more views, making this method fast.

%% file: 04d-DetailedAlgorithm.tex
We now describe our algorithm for estimating a single landmark triplet observed from multiple views using noisy measurements and a motion model. Usually, SLAM solvers use a camera-centric global frame. We use a different, landmark-centric frame by fixing landmarks $A=(0,0)$ and $B=(0,1)$, as we believe it is more intuitive to infer qualitative landmark-related states (Figure  \ref{fig:ABframe}). \cite{pierlot2014tro} gives a complete and efficient solution to the \emph{deterministic} problem.

First, we introduce a few notations to help describe the algorithm. While the random variable $X_n$ represents camera pose at time $n$, we denote \emph{samples} of this camera pose as $X^{(k_n)}_n$, where $n$ and $k_n$ are, respectively, time and sample indices. We also define a "trajectory hypothesis" $th^j_n \doteq \{X^{(k_1)}_1, \dots, X^{(k_n)}_n\}$ as a set of specific camera pose samples - one for each time step. Index $j \mapsto \{k_1, \ldots,k_n\}$ is a simplified notation for a trajectory hypothesis. With a slight abuse of notation, we denote $X^j_l$ as the camera pose sample from time $l$ in trajectory hypothesis $th^j_n$. The set of all trajectory hypotheses at time $n$ is $TH_n \doteq \{th^j_n\}$.

To estimate the qualitative state of a landmark triplet with measurements from $n$ time steps, we iteratively apply a three-stage algorithm for each time step:

\emph{Sampling step: } Generate $m_n$ camera pose samples for time $n$ $X^{(k_n)}_n$, with $k_n\in [1,m_n]$. We samples from the distribution $\prob{X_n| \phi_n^A, \phi_n^B}$ using bearing  measurements  $\phi_n^A$ and $\phi_n^B$ to landmarks $A$ and $B$. Given these measurements, the 2D camera pose is on a specific part of a \emph{circle} that goes through the two landmarks A and B (see Figure \ref{fig:2pResecsubfig1}).  \cite{pierlot2014tro} specifies the calculation of the \emph{locus circle} parameters. Camera poses are sampled in the vicinity of this locus circle considering the noisy nature of $\phi_n^A$ and $\phi_n^B$.

\emph{Motion step: } For each  trajectory hypothesis $th^j_{n-1} \in TH_{n-1}$, we can use $X^j_{n-1}$ and motion azimuth $\psi_{n-1}$ to intersect the locus circle from time $n$ 
(see Figure \ref{fig:2pResecsubfig2}). Camera pose samples $X^{(k_n)}_n$ that are consistent with this intersection are found, also considering the noisy nature of the motion model \eqref{eq:MotionModel}. Using these matches, we generate multiple new, extended trajectory hypotheses $th^j_n$. For each $th^j_n$, we also calculate and keep motion model consistency wight: $wm^j_n=\prob{X^{j}_n | X^j_{n-1}, \psi_{n-1}}$.

\emph{Resection step: } For each valid trajectory hypothesis $th_n^j$, we test the consistency of the bearing measurements from all cameras to landmark C. First, we use camera poses $X^j_{1:n}$ and bearing measurements $\phi^{C}_{1:n}$ to triangulate landmark C location, denoted by $L^{C,j}$. It is estimated to be the centroid of all line of sight pairs intersection points (see Figure \ref{fig:2pResecsubfig2}). Then we estimate a triangulation weight as the probability for this configuration: $wr^j_i=\prob{\phi^C_i|L^{C,j}, X^j_i} , \forall i\in \{1...n\}$ (assuming independent measurement noise). We disqualify trajectories that do not intersect or have low probability.

Thus far, the results of our algorithm are the set of valid trajectory hypotheses $th^j_n \in TH_n$. For each , $th^j_n$ we keep a single landmark location hypothesis $L^{C,j}$ along with corresponding consistency weights $wm^j_i$ and triangulation weights $wr^j_i$  $\forall i\in \{1...n\}$.

\emph{Note:} Using a single $L^{C,j}$ for each  $th^j_n$ is a heuristic. We can sample over the area of line-of-sight intersections to cover all probable $L^C$ locations. We choose the simpler way for run-time considerations.

\emph{Note:} If we have measurements from only one time step, we sample landmark C location $L^C$ around each sampled camera pose $X_1^{(k1)}$ line of site. We consider measurement noise and sample from $\prob{L^C|X_1^{(k1)},\phi_1^C}$.

Finally, we approximate Eqs.~\eqref{eq:slam1} and \eqref{eq:slam2} by summing over each qualitative state to get state probability distribution (see the similarity to \eqref{eq:slamst3}):

\begin{equation}\label{eq:slamapprox1}
	\prob{s_k^C | H_n} \approx \eta^C \sum\displaylimits_{j}{ \mathds{1}(L^{C,j} \in s_k^C) wr_1 \prod_{i=2}^n wm^j_i wr^j_i },
\end{equation}
\begin{equation}\label{eq:slamapprox2}
	\prob{ s_k^{X_i} | H_n} \approx \eta^{X_i} \sum\displaylimits_{j}{\mathds{1}(X^j_i \in s_k^{X_i}) wr_1 \prod_{i=2}^n wm^j_i wr^j_i }.
\end{equation}
Here, $\eta^C$ and $\eta^{X_i}$ are normalization constants, and the sum is over all trajectory hypotheses $th_n^j\in TH_n$. Note the term $wr_1 \prod_{i=2}^n wm^j_i wr^j_i$ is a sampled approximation of the joint pdf $\prob{X_{1:n},L_C| H_n}$. 

This algorithm seems to handle many trajectory hypotheses that grow exponentially in time. Practically, the geometric constraints enforced in the "motion" step and the "resection" step dramatically decrease the number of hypotheses for two views and even more for three views or more, making our algorithm much faster (experiment example in figure \ref{fig:res_example}). This reduction is an effect of incorporating a motion model into the qualitative estimation. It also means that using this algorithm incrementally requires saving only a small number of hypotheses and therefore is not memory intensive. A pseudo-code for a simplified version of this approach is in Algorithm \ref{alg:alg1}.

\begin{algorithm}
	\caption{Single triplet qualitative state estimation}\label{alg:alg1}
	\footnotesize
	\begin{algorithmic}[1]
		\State  sample camera poses $X^{(k_1)}_1 \sim \prob{X_1|\;\phi_1^A, \,\phi_1^B}$ with $k_1\in [1, m_1]$
		\State  initialize trajectory hypothesis set: $TH=\{X^{(k_1)}_1\}$
		\For{$i =1,...,n$}
		\State {\color{blue} // sampling step:}
		\State sample camera pose $X^{(k_i)}_i \sim \prob{X_i|\;\phi_i^A,\, \phi_i^B}$ with $k_i=[1, m_i]$
		\State {\color{blue} // motion model step:}
		\State    -  $\forall th^j_{i-1}$ find $X^j_i$ that are consistent with motion azimuth $\psi_{i-1}$
		\State   -   extend $TH_i$ with new matches
		\State   -  keep consistency weight: $wm_{i}^j=\prob{X^{(k_i)}_i |\; X^j_{i-1},\, \psi_{i-1}}$
		\State {\color{blue} // resection step:}
		\State    -   $\forall\; th^j_i$ find $L^{C,j}$ by line of sight triangulation
		\State    -   keep consistency weight: $wr_{t}^j=\prob{\phi^C_t|\;L^{C,j},\, X^j_t} ,\, \forall\, t\in \{1...i\}$
		\State    -   dismiss low probability hypotheses
		\EndFor\label{euclidendwhile}
		\State estimate landmark qualitative state probability via Eq.~\eqref{eq:slamapprox1}
		\State estimate camera qualitative state probability via Eq.~\eqref{eq:slamapprox2}
	\end{algorithmic}
\end{algorithm}

\emph{Remark:} This algorithm is simplified for the sake of explanation. Our actual implementation is more optimized.

%% file: 04e-FasterVariantAlgorithm.tex
We seek to utilize further the ability of the qualitative coarse resolution to absorb errors, aiming to achieve an even faster algorithm. With this motivation in mind, we introduce an approach similar to the sampling-based approach described above. However, sampling is done only to represent the prior term in Eq.~\eqref{eq:slamst3}. No sampling is done to represent the motion model or measurement noise. 

In Section \ref{sec:SingleTripletResults}, we evaluate this algorithm variant to see how many noise-related errors it can handle. As we shall see, due to the inherent properties of qualitative estimation, it can handle reasonable noise levels with accuracy similar to the full algorithm and with a significant speedup.

%% file: 05-approach_composition.tex
\vspace{5pt}

Up until now, we discussed how to solve a single landmark triplet problem. We showed how to estimate the qualitative state of the camera and landmarks in landmark relative frames. 

The second part of our work addresses the multiple landmark qualitative map and how to propagate data between different landmark triplets in this map. We start with the basic concept of  \emph{qualitative composition} - an inherent geometric attribute of landmark triplets with partially common landmarks to transfer qualitative data between them (section \ref{sec:Composition}). Qualitative data propagation was first approached in the seminal work \cite{Mcclelland14jais} but in a basic non-probabilistic sense. We are the first ones (as far as we know) to formulate probabilistic composition. We also formulate the connection between basic qualitative composition inference to the full solution of the problem of propagating information in a qualitative map given prior data. Then in chapter \ref{sec:CompositionFG} we discuss the representation of a qualitative map as a factor graph. Finally, in section \ref{sec:FactorGraphPropagation}, we present a study of how composition propagates data in the qualitative factor graph regarding graph topology and prior knowledge, including novel information metrics and an algorithm for graph propagation we developed for this end.

\subsection{Propagating Data Between Triplets - Probabilistic Composition}
\label{sec:Composition}
\input{05a-Composition}

\subsection{Qualitative Map As a Factor Graph}
\label{sec:CompositionFG}
\input{05b-CompositionFG}

\subsection{A Study Of Factor Graph Propagation By Composition}
\label{sec:FactorGraphPropagation}
\input{05c-FactorGraphPropagation}

%% file: 05a-Composition.tex
\begin{figure}[]
	\centering
	\subfloat[]{\includegraphics[width=0.26\textwidth]{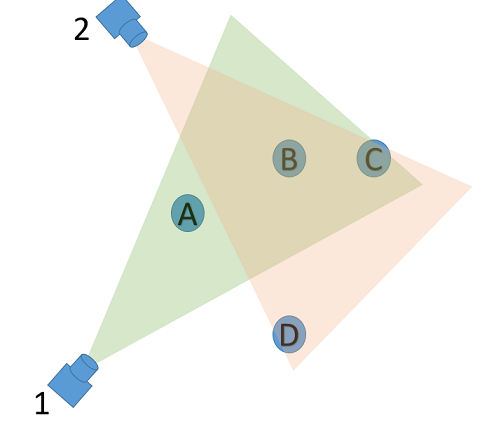}\label{fig:map1} }\quad
	\subfloat[]{\includegraphics[width=0.26\textwidth]{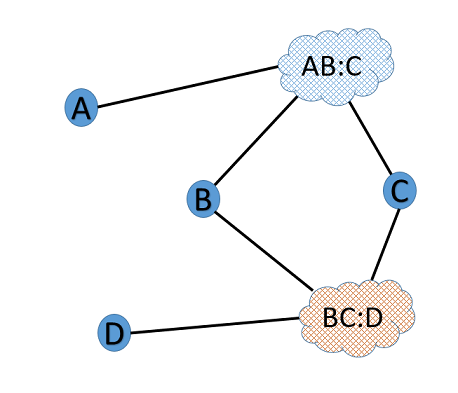}\label{fig:map2} }\quad	
	\subfloat[]{\includegraphics[width=0.31\textwidth]{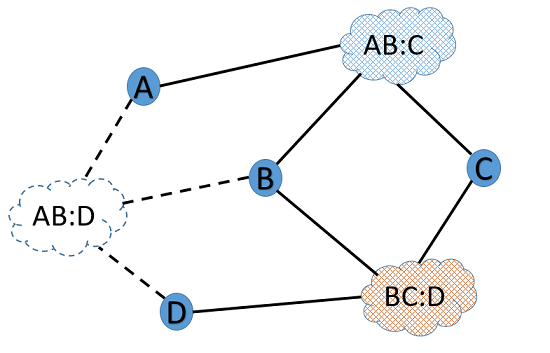}\label{fig:composition} }\\
	\caption{\label{fig:qualittiveMap}qualitative map:  (a) Landmarks A,B,C observed from camera pose 1. Landmarks B,C,D observed from camera pose 2. (b)qualitative map is represented as landmark triplet graph (c) Composition: try to estimate triplet AB:D given only AB:C and BC:D estimations (for unobserved triplets, or improving existing estimations). \vspace{-12pt}}
\end{figure}

In \cite{Mcclelland14jais}, it is noted that if three landmark triplets share two common landmarks between each pair (e.g., AB:C, BC:D, AB:D), there are intrinsic geometric constraints about the joint feasibility of the three qualitative states $(s^{AB:C}_i,s^{BC:D}_j,s^{AB:D}_k)$ (without referring to any prior knowledge). Applying these constraints to infer the joint qualitative state is called  \emph{composition}. Composition enables enhanced estimation for existing overlapping triplets and to infer triplets never viewed directly (see Figure \ref{fig:qualittiveMap}). Among other things, this is an essential basis for managing large-scale qualitative maps and active qualitative planning. 

However, \cite{Mcclelland14jais} only refers to the non-probabilistic joint feasibility of qualitative EDC states. Also, \cite{Mcclelland14jais} does not formulate the general problem of propagating data between the three triplets given measurement history or show how composition relates to it. In this section, we develop a probabilistic composition variant and show how to use it for data propagation. We also formulate the general problem of inferring the joint qualitative states probability given measurements $\prob{S^{AB:C},S^{BC:D},S^{AB:D}|H_n^{AB:C},H_n^{BC:D},H_n^{AB:D}}$. Then we show how composition-based inference relates to the full solution.

For the sake of readability, we simplify notations in this section. We drop the time index $n$ and assume all history is considered. We also use shortened notations for triplets, as shown in section \ref{sec:problem-formulation}. AB:C, BC:D, AB:D are denoted as $t1, t2$ and $t3$ respectively. So for example: qualitative states is denoted $S^{AB:C}\equiv S^{t1}$, history is denoted $H^{AB:C}\equiv H^{t1}$, and the landmark metric location is denoted  $L^{AB:C}\equiv L^{t1}$.

We address inferring joint qualitative probability $\prob{S^{t1},S^{t2},S^{t3}|H_n^{t1},H_n^{t2},H_n^{t3}}$, where triplets  AB:C, BC:D, AB:D are denoted as $t1, t2$ and $t3$ respectively. Specifically, this formulation is also valid in case some of the triplets have no observations, e.g., $\prob{S^{t1},S^{t2},S^{t3}|H_n^{t2},H_n^{t3}}$.

For the same reasons as in section \ref{sec:Probabilistic navigation and mapping}, we consider the underlying 4-landmark $(A,B,C,D)$ metric problem. We look at the joint probability of the three triplets and marginalize the metric locations for each triplet:
\begin{equation*}
	\!\begin{multlined}[t]
			\prob{S^{t1},S^{t2},S^{t3}|H^{t1},H^{t2},H^{t3}}
			=\iiint\displaylimits_{L^{p1},L^{p2},L^{t}}{ \prob{S^{t1},S^{t2},S^{t3},L^{t1},L^{t2},L^{t3}| H^{t1}, H^{t2}, H^{t3}} dL^{t1}dL^{t2} dL^{t3}}.
	\end{multlined}
\end{equation*}
We break the integration argument using the formula of total probability:
\begin{equation*}
	\!\begin{multlined}[t]
		\prob{S^{t1},S^{t2},S^{t3},L^{t1},L^{t2},L^{t3}| H^{t1}, H^{t2}, H^{t3}}
		= \prob{S^{t1},S^{t2},S^{t3}|L^{t1},L^{t2},L^{t3},H^{t1},H^{t2},H^{t3}} \prob{L^{t1},L^{t2},L^{t3}| H^{t1}, H^{t2}, H^{t3}}.
	\end{multlined}
\end{equation*}	
Note that given the metric location of the landmarks $L^{t1},L^{t2},L^{t3}$, the qualitative states $S^{t1},S^{t2},S^{t3}$ are directly determined, and are independent in anything else. Therefore we get the following:
\begin{equation*}
	\!\begin{multlined}[t]
		\prob{S^{t1},S^{t2},S^{t3},L^{t1},L^{t2},L^{t3}| H^{t1}, H^{t2}, H^{t3}}=
		= \prob{S^{t1},S^{t2},S^{t3}|L^{t1},L^{t2},L^{t3}}		\prob{L^{t1},L^{t2},L^{t3}| H^{t1}, H^{t2}, H^{t3}},
	\end{multlined}
\end{equation*}
and overall:
\begin{equation} \label{eq:metric_propagation}
	\!\begin{multlined}[t]
		\prob{S^{t1},S^{t2},S^{t3}|H^{t1},H^{t2},H^{t3}}= 
		\!\!\!\!\!\!\iiint\displaylimits_{L^{t1},L^{t2},L^{t3}}{\prob{S^{t1},S^{t2},S^{t3}|L^{t1},L^{t2},L^{t3}} \prob{L^{t1},L^{t2},L^{t3}| H^{t1}, H^{t2}, H^{t3}} }
		  dL^{t1} dL^{t2} dL^{t3}.
	\end{multlined}
\end{equation}
This formulation resembles equation \eqref{eq:slam1}. It gives a similar intuitive result: For each combination of qualitative states $S^{t1},S^{t2},S^{t3}$, we solve the metric 4-landmark SLAM problem $\prob{L^{t1},L^{t2},L^{t3}| H^{t1}, H^{t2}, H^{t3}}$, and integrate probability distribution over all landmark location that corresponds with this combination of qualitative states.

However, using this method in a sizeable multiple-landmark map will eventually translate (when using variable elimination) to solving the big multi-landmark SLAM metric problem.  We want to use qualitative geometry's unique properties to find an approximate faster and more straightforward solution and still achieve good results. Specifically, we want to replace the second factor in equation \eqref{eq:metric_propagation} with a probabilistic version of the composition described in \cite{Mcclelland14jais}.

Our main approximation is as follows: 
\begin{itemize}
	\item Instead of considering the full 4-landmark$(A,B,C,D)$ problem, we estimate each triplet's qualitative state using only its own history.
	\item When propagating information between different triplets, we use only the estimated qualitative state of the triplets instead of the entire history.
\end{itemize}
This approximation might be considered a blunt for metric SLAM. However, we will show that since we care only about the qualitative states, it is much less influential and still enables considerable data propagation in the qualitative map. On the other hand, it gives us several advantages looking at the overall approach:
\begin{enumerate}
	\item Each triplet can be estimated individually using the incremental algorithm in Section \ref{sec:ApproachSingleTriplet}.
	\item Map propagation uses only the qualitative state for each triplet. So in a large map, triplets can be saved efficiently using very little data (if no intermediate estimation results are saved for incremental algorithms).
	\item We will show that the composition-based algorithm is straightforward and fast and can be mostly calculated offline. Low computing is significant for online map propagation since every new measurement propagates multiple times through the map in each time step.
\end{enumerate}

To apply these approximations for the second factor in equation \eqref{eq:metric_propagation}, we start by marginalizing over the qualitative states:
\begin{equation*}
	\!\begin{multlined}[t]
		\prob{L^{t1},L^{t2},L^{t3}| H^{t1}, H^{t2}, H^{t3}}
		=\sum_{s^{t1}_i}\sum_{s^{t2}_j}\sum_{s^{t3}_k}{\prob{L^{t1},L^{t2},L^{t3},s^{t1}_i,s^{t2}_j,s^{t3}_k| H^{t1}, H^{t2}, H^{t3}}}.
	\end{multlined}
\end{equation*}
Using the formula of total probability to break down the sum argument we get:
\begin{equation*}
	\!\begin{multlined}[t]
		\prob{L^{t1},L^{t2},L^{t3}| H^{t1}, H^{t2}, H^{t3}}
		=\sum_{s^{t1}_i}\sum_{s^{t2}_j}\sum_{s^{t3}_k}{\prob{L^{t1},L^{t2},L^{t3}|s^{t1}_i,s^{t2}_j,s^{t3}_k, H^{t1}, H^{t2}, H^{t3}}} 
		\prob{s^{t1}_i,s^{t2}_j,s^{t3}_k| H^{t1}, H^{t2}, H^{t3}}.
	\end{multlined}
\end{equation*}
The first part of our approximation uses only the qualitative states to estimate landmark metric locations instead of the entire history:
\begin{equation*}
	\!\begin{multlined}[t]
		\prob{L^{t1},L^{t2},L^{t3}| H^{t1}, H^{t2}, H^{t3}}
		\approx \sum_{s^{t1}_i}\sum_{s^{t2}_j}\sum_{s^{t3}_k}{\prob{L^{t1},L^{t2},L^{t3}|s^{t1}_i,s^{t2}_j,s^{t3}_k}
		\prob{s^{t1}_i,s^{t2}_j,s^{t3}_k| H^{t1}, H^{t2}, H^{t3}}}.
	\end{multlined}
\end{equation*}
The second part of our approximation assumes that each qualitative state is estimated only based on its own history and correlations between different triplets are dropped:
\begin{equation*}
	\!\begin{multlined}[t]
		\prob{L^{t1},L^{t2},L^{t3}| H^{t1}, H^{t2}, H^{t3}}
		 \approx \sum_{s^{t1}_i}\sum_{s^{t2}_j}\sum_{s^{t3}_k}{\prob{L^{t1},L^{t2},L^{t3}|s^{t1}_i,s^{t2}_j,s^{t3}_k}
		\prob{s^{t1}_i| H^{t1}} \prob{s^{t2}_j| H^{t2}} \prob{s^{t3}_k|  H^{t3}}}.
	\end{multlined}
\end{equation*}
We get the approximated $\tilde{\mathbb{P}}({S^{t1},S^{t2},S^{t3}|H^{t1},H^{t2},H^{t3}})$ By substituting this back to equation \eqref{eq:metric_propagation}: 
\begin{equation*}
	\!\begin{multlined}[t]
		\prob{S^{t1},S^{t2},S^{t3}|H^{t1},H^{t2},H^{t3}}
		\approx \iiint\displaylimits_{L^{t1},L^{t2},,L^{t3}}{\prob{S^{t1},S^{t2},S^{t3}|L^{t1},L^{t2},L^{t3}}  }
		\sum_{s^{t1}_i}\sum_{s^{t2}_j}\sum_{s^{t3}_k}{\prob{L^{t1},L^{t2},L^{t3}|s^{t1}_i,s^{t2}_j,s^{t3}_k}}\\
		\prob{s^{t1}_i| H^{t1}} \prob{s^{t2}_j| H^{t2}} \prob{s^{t3}_k|  H^{t3}}  dL^{t1} dL^{t2} dL^{t3} 
		\triangleq \tilde{\mathbb{P}}({S^{t1},S^{t2},S^{t3}|H^{t1},H^{t2},H^{t3}}).
	\end{multlined}
\end{equation*}
Now we change the order of integration and sum:
\begin{equation*}
	\!\begin{multlined}[t]
		\tilde{\mathbb{P}}({S^{t1},S^{t2},S^{t3}|H^{t1},H^{t2},H^{t3}}) = 
		\sum_{s^{t1}_i}\sum_{s^{t2}_j}\sum_{s^{t3}_k}{\prob{s^{t1}_i| H^{t1}} \prob{s^{t2}_j| H^{t2}} \prob{s^{t3}_k|  H^{t3}}} \\   \iiint\displaylimits_{L^{t1},L^{t2},L^{t3}}{\prob{S^{t1},S^{t2},S^{t3}|L^{t1},L^{t2},L^{t3}}  }
		\prob{L^{t1},L^{t2},L^{t3}|s^{t1}_i,s^{t2}_j,s^{t3}_k}  dL^{t1} dL^{t2} dL^{t3},
	\end{multlined}
\end{equation*}
and we note that
\begin{equation}
	\!\begin{multlined}[t]
	\prob{S^{t1},S^{t2},S^{t3}|L^{t1},L^{t2},L^{t3}}=
	\begin{cases}
		1, & \text{if}\ L^{ti} \in S^{ti} \quad \forall i=1,2,3 \\
		0, & \text{otherwise}
	\end{cases}.
	\end{multlined}
\end{equation}
Therefore $\prob{S^{t1},S^{t2},S^{t3}|L^{t1},L^{t2},L^{t3}}$ can be translated into integral bounds:
\begin{equation*} \label{eq:probabilisticComposition}
	\!\begin{multlined}[t]
		\tilde{\mathbb{P}}({S^{t1},S^{t2},S^{t3}|H^{t1},H^{t2},H^{t3}}) = \\
		= \sum_{s^{t1}_i}\sum_{s^{t2}_j}\sum_{s^{t3}_k}{\prob{s^{t1}_i| H^{t1}} \prob{s^{t2}_j| H^{t2}} \prob{s^{t3}_k|  H^{t3}}}
		 \iiint\displaylimits_{L^{t1}\in S^{t1} ,L^{t2}\in S^{t2},L^{t3}\in S^{t3}}{\prob{L^{t1},L^{t2},L^{t3}|s^{t1}_i,s^{t2}_j,s^{t3}_k}  }
		dL^{t1} dL^{t2} dL^{t3}.
	\end{multlined}
\end{equation*}

This approximate formulation is very enlightening. The first term is the stand-alone estimation of each triplet separately: $\prob{s^{t1}_i| H^{t1}} \prob{s^{t2}_j| H^{t2}} \prob{s^{t3}_k|  H^{t3}}$  as detailed in section \ref{sec:Probabilistic navigation and mapping} equation \ref{eq:slam1}. The second term is \emph{qualitative composition}. It is an intrinsic geometric constraint between the triplets independent of measurements. It represents the probability of each specific combination of qualitative states: $ \{ s^{t1}_i, s^{t2}_j, s^{t3}_k \} \forall i,j,k\in\{1...m\}$. Therefore it can also be calculated \emph{offline} for each of the $m^3$ combinations of the three qualitative states.

\begin{figure}[]
	\centering
	
	\subfloat[]{\includegraphics[width=0.30\textwidth]{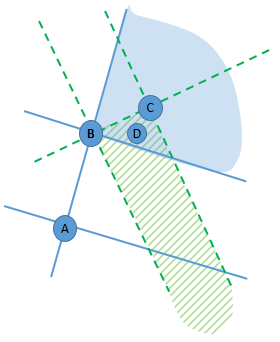}\label{fig:cprob1} }\quad
	\subfloat[]{\includegraphics[width=0.30\textwidth]{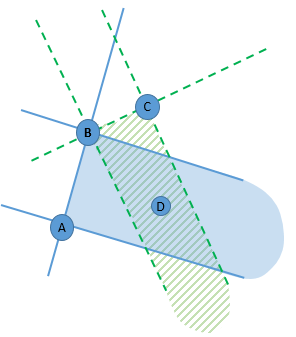}\label{fig:cprob2} }\quad
	\subfloat[]{\includegraphics[width=0.30\textwidth]{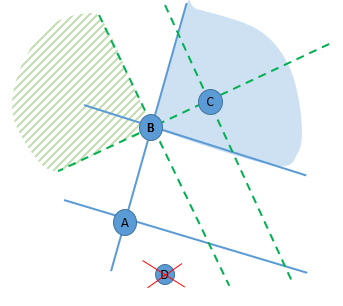}\label{fig:cprob3} }\quad
	\subfloat[]{\includegraphics[width=0.30\textwidth]{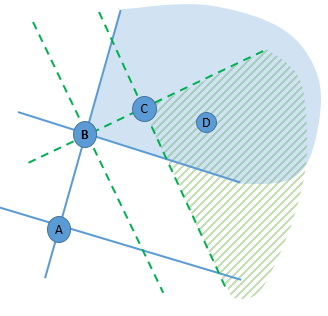}\label{fig:cprob4} }\quad
	\subfloat[]{\includegraphics[width=0.2\textwidth]{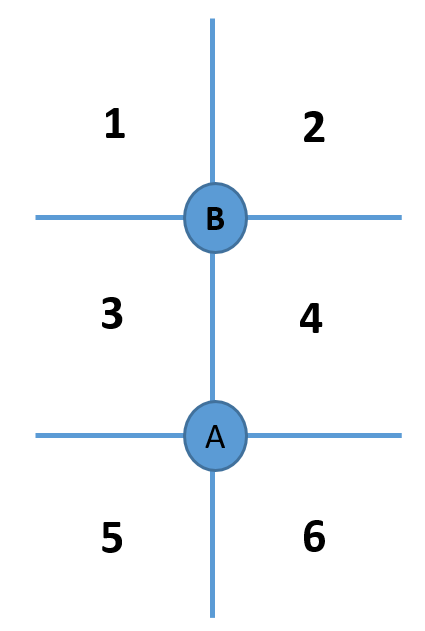}\label{fig:doubleCrossNum} }\\
	\caption{\label{fig:compositionProbablitiy} Example for our implementation of qualitative composition offline calculation in section \ref{sec:Composition}. This example uses Freska's double cross space partition (e). For each specific combination of qualitative states  $(s^{AB:C}_i, s^{BC:D}_j, s^{AB:D}_k)$ we sample $L^{AB:C} \in s^{AB:C}_i$ uniformly. Given each sample, we calculate the intersection area of the polygons corresponding to $s^{BC:D}_j$ and $s^{AB:D}_k$ in the $AB$ frame. We sum for all $L^{AB:C}$ samples and normalize. Figures (a-d) demonstrate a specific sample $L^{AB:C} \in S^{AB:C}=2$ and different $s^{BC:D}_j$ and $s^{AB:D}_k$ combinations: (a) small intersection for $S^{BC:D}=4$ and $S^{AB:D}=2$. (b) larger intersection for $S^{BC:D}=4$ and $S^{AB:D}=4$. (c) No intersection for $S^{BC:D}=5$ and $S^{AB:D}=2$. (d) Infinite intersection for $S^{BC:D}=2$ and $S^{AB:D}=2$ requires special treatment.}
\end{figure}

We implement this \emph{offline} calculation using a semi-sample-based approach. For each qualitative state combination $(s^{AB:C}_i, s^{BC:D}_j, s^{AB:D}_k)$ we sample $L^{AB:C}$ from $s^{AB:C}_i$ uniformly. Then we calculate the intersection area of the geometric polygons that fit $s^{BC:D}_j \cap s^{AB:D}_k$ in the $AB$ frame (for a specific qualitative space partition). Finally, we sum the intersection areas for all samples and normalize the probability for all $(s^{AB:C}_i, s^{BC:D}_j, s^{AB:D}_k)$ combinations to sum up to 1. This process is illustrated in figure \ref{fig:compositionProbablitiy} using  Freska's double cross space partition \cite{Freksa92smc}. In \ref{fig:cprob1},\ref{fig:cprob2} we see two different $(s^{BC:D}_j, s^{AB:D}_k)$ combinations given a specific sample of $L^{AB:C} \in s^{AB:C}_i$, each with a different intersection area. Some combinations of $(s^{BC:D}_j, s^{AB:D}_k)$ are infeasible given a specific sample of $L^{AB:C} \in s^{AB:C}_i$ and have $0$ intersection area, and therefore $0$ probability (e.g. figure \ref{fig:cprob3}). On the other hand, some combinations have an infinite intersection area (e.g., figure \ref{fig:cprob4}). If not handled correctly, this could zero out all finite intersection combinations. Therefore we enforce integration borders for each landmark triplet relative frame. We choose these borders to achieve as similar area size for all states as possible and thereby minimize bias (different for each qualitative space partition). As an example, for  Freska's double cross \cite{Freksa92smc} we use $-1<x<1, -1<y<2$. The deterministic qualitative composition in \cite{Mcclelland14jais} considers only the feasibility of qualitative state combinations. This is equivalent to reducing the probabilities for each  $(s^{AB:C}_i, s^{BC:D}_j, s^{AB:D}_k)$ combination to $1$ or $0$ (and normalize).

The online calculation of equation \ref{eq:probabilisticComposition} can now be very efficient. This efficiency is more clearly demonstrated by using vector and matrix notations:
\begin{equation*}
	\begin{split}
		& V^{ti}_{m \times 1} \triangleq \prob{S^{ti}|H^{ti}}, \\
		& \!\begin{multlined}[t]
			 T^{t1,t2,t3}_{m \times m \times m}(i,j,k) \triangleq 
			  \iiint\displaylimits_{L^{t1}\in S^{t1} ,L^{t2}\in S^{t2},L^{t3}\in S^{t3}}{\prob{L^{t1},L^{t2},L^{t3}|s^{t1}_i,s^{t2}_j,s^{t3}_k}  }
			   dL^{t1} dL^{t2} dL^{t3}.
			\end{multlined}
	\end{split}
\end{equation*}

Here $\prob{S^{t1}| H^{t1}}, \prob{S^{t2}| H^{t2}}, \prob{S^{t3}| H^{t3}}$ are denoted as a $m \times 1$ vectors, and the offline integral term in Eq. \ref{eq:probabilisticComposition} for  each of the $m^3$ combination of qualitative states $ \{ s^{t1}_i, s^{t2}_j, s^{t3}_k \}$ is denoted as a $m \times m \times m$ tensor. We also note that it is sparse to some degree, e.g.~offline composition for EDC space partition is a $T^{t1,t2,t3}_{20 \times 20 \times 20}$ tensor with 8000 elements in total, but only 2257 non-zero elements.

For example, if we want the probability for a specific combination of qualitative states, equation \ref{eq:probabilisticComposition} translates to: 
\begin{equation} \label{eq:specificComposition}
	\!\begin{multlined}[t]
		\tilde{\mathbb{P}}({s^{t1}_i,s^{t2}_j,s^{t3}_k|H^{t1},H^{t2},H^{t3}}) =
		 V^{t1}(i) V^{t2}(j)  V^{t3}(k) T(i,j,k),
	\end{multlined}
\end{equation}
and if we want to calculate the marginal for a specific triplet, equation \ref{eq:probabilisticComposition} translates to: 
\begin{equation}\label{eq:marginalComposition}
	\!\begin{multlined}[t]
	\tilde{\mathbb{P}}({S^{t1}|H^{t1},H^{t2},H^{t3}})
	=\sum_{s^{t2}_i}{\sum_{s^{t3}_j}{\prob{S^{t1},S^{t2},S^{t3}|H^{t},H^{p1},H^{p2}} }} 
	= (T_{m \times m \times m} \times V^{t2}_{m \times 1}) \times V^{t3}_{m \times 1}.
	\end{multlined}
\end{equation}
This is a speedy way to update joint probabilities and marginals since it uses only (and not too many) mult-add operations. We will also see in sections \ref{sec:BasicCompositionFactorAnalysis} and  \ref{sec:FactorGraphPropagationAnalysis} that although the composition is an approximation, it is effective and still propagates significant amounts of qualitative data. Thus our primary motivation to enable online data propagation in the qualitative map for low-compute platforms is supported.
	
To summarize, in this section, we formulated the probabilistic qualitative composition for the first time and discussed the exact meaning of the approximation concerning the full solution for propagating data in the qualitative map. Note that this formulation of qualitative composition is a generalized version of the formulation from the early conference version of this work \cite{Mor20iros}. We give here the full joint probability  $\prob{S^{t1},S^{t2},S^{t3}|H^{t1},H^{t2},H^{t3}}$, instead of the marginal $\prob{S^{t1}|H^{t2},H^{t3}}$, and discuss the meaning of the approximation and implementation in more detail.

%% file: 05b-CompositionFG.tex
We discussed the basic geometry of propagating data between landmark triplets and a fast approximate algorithm. We will show how to use this in a factor graph representation of the multi-landmark qualitative map to maintain and propagate data.

\begin{figure}[]
	\centering
	\includegraphics[width=0.5\textwidth]{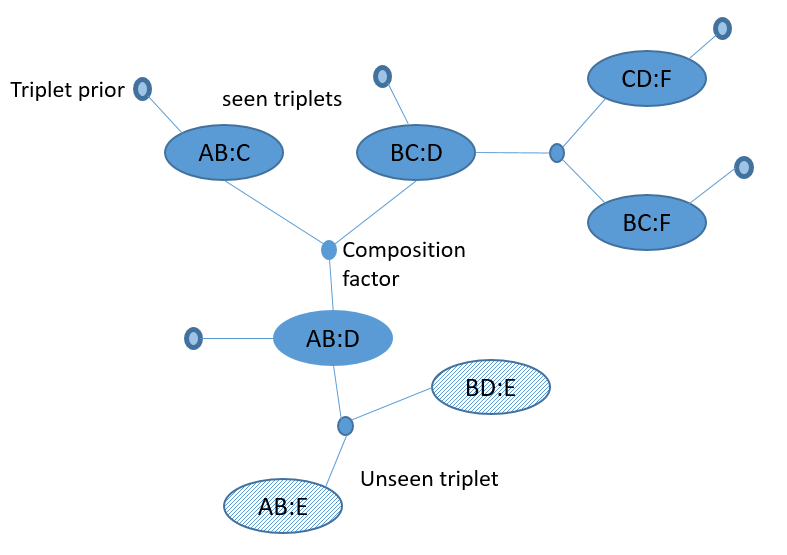}\label{fig:fg1} 
	\caption{\label{fig:qualitative_factor_graph}Qualitative map as a factor graph: Variable nodes are landmark triplet qualitative states. Prior factor nodes - estimating landmark triplet qualitative state based on its measurements. Composition factor nodes - connecting between three triplets with common landmarks by composition.  }
\end{figure}

Factor graphs are used in state-of-the-art SLAM (and active planning) approaches as discussed in detail, e.g.~in \cite{dellaert2017factor}.  Generally speaking, a factor graph is a bipartite graph $G$ that contains two types of nodes. Variable nodes $V_v$ represent the random variables of the problem, and factor nodes $V_f$ encode probabilistic constraints between different variables. The edges in the graph $E$ connect each factor node to all the variable nodes it engages (see Fig.~\ref{fig:qualitative_factor_graph}):
\begin{equation*}
	\begin{split}
		& G=(V,E)\\
		& V = V_v\cup V_f.
	\end{split}
\end{equation*}
In our framework, the qualitative map is the collection of the qualitative states of all landmark triplets we are interested in $M=\{S^{i,j:k}\}$. In the current formulation, we do not include camera poses as part of the map for simplicity. The unknown variable nodes, in this case, are the corresponding qualitative states:
\begin{equation*}
	V_v=\{ v_v^{ij:k}\} \quad \forall S^{i,j:k}\in M.
\end{equation*}
These landmark triplets can be those we have seen together and measured in the past, denoted as  $V_v^{seen}$, or unseen triplets $V_v^{unseen}$ that we are interested in for accomplishing a specific task or mission such as planning (see  Fig.~\ref{fig:qualitative_factor_graph}). Note that unseen triplets mean the landmarks were not observed together from the same view but might have been observed separately from different views.
\begin{equation*}
	\begin{split}
		& V_v=V_v^{seen}\cup V_v^{unseen}\\
		& V_v^{seen}=\{ v_v^{ij:k} \} \quad \forall S^{i,j:k}\in M , H^{ij:k}\neq \emptyset\\
		& V_v^{unseen}=\{ v_v^{ij:k} \} \quad \forall S^{i,j:k}\in M ,  H^{ij:k}=\emptyset.
	\end{split}
\end{equation*}
We also discuss two types of factors: First is the estimation of each landmark triplet qualitative state by its own measurement history $\prob{S^{ij:k}|H^{ij:k}_n}$ as calculated in section \ref{sec:Probabilistic navigation and mapping}. This factor is unary (connected only to one variable node) and exists for any landmark triplet $V_v^{seen} \in V^{seen}$ with observations. The group of all unary factors is denoted as $V_{fu}$. The second factor is a \emph{"composition factor"}. It is a trinary factor that involves landmark triplets with common landmarks connected by composition, as discussed in Section \ref{sec:Composition}. The group of all composition factors is denoted as $V_{fc}$. Therefore we get the following:
\begin{equation*}
	\begin{split}
		& V_f=V_{fu}\cup V_{fc}\\
		& V_{uf} = \{ v_{fu}^{ij:k}\} \quad \forall \; v_v^{ij:k} \in  V_v^{seen} \\
		& V_{cf} = \{ v_{fc}^{i,j,k,s}\} \quad \forall\; i,\,j,\,k,\,s ;\; S^{i,j:k},\,S^{j,k:s},\,S^{i,j:s}\in M.
	\end{split}
\end{equation*}
Factor graph edges correspondingly are of two types. \emph{Unary edges} exist between every unary factor and its corresponding triplet qualitative state, and \emph{composition edges} exist between any composition factor to the three common landmark triplets it is associated with:
\begin{equation*}
	\begin{split}
		& E = E_u\cup E_c\\
		& E_u = \{(v_v^{ij:k}, v_{uf}^{ij:k})\} \quad \forall  v_v^{ij:k} \in  V_v^{seen} \\
		& \!\begin{multlined}[t]
			E_c = \{(v_{fc}^{i,j,k,s},v_v^{ij:k}),(v_{fc}^{i,j,k,s},v_v^{jk:s}),(v_{fc}^{i,j,k,s},v_v^{ij:s})\} 
     		\forall v_{fc}^{i,j,k,s}\in V_{cf}.
     	\end{multlined}
	\end{split}
\end{equation*}
It is important to note that our factor graph represents a joint distribution over \emph{discrete} random variables. As mentioned in section \ref{sec:Composition}, variable nodes and unary factor nodes are $m\times 1$ vectors, whereas composition nodes are $m\times m\times m$ tensors. Fig.~\ref{fig:qualitative_factor_graph} is a visual example of this qualitative factor graph.

%% file: 05c-FactorGraphPropagation.tex
Having defined our qualitative SLAM framework and how to represent it as a qualitative factor graph, we now study how data propagation by composition behaves in the factor graph. The main goal is to see how much information can composition propagate through the graph in relation to prior knowledge and graph topology. This study is also essential for addressing active qualitative planning (specifically the ability to estimate unseen triplets). We only perform a basic study and do not suggest a generally optimal solution for qualitative factor graph data propagation (this can be thoroughly done in future work).

This study directly relates to topological signatures in traditional metric SLAM factor graphs with Gaussian distributions. The works \cite{Khosoussi18ijrr, Kitanov18icra, Kitanov19arxiv, Shienman21ral} show that topological graph signatures are highly correlated to differential entropy (under certain assumptions). We provide the first empirical indication that a similar notion could potentially apply also to factor graphs with discrete variables.

The simplest way to measure data propagation in the factor graph is by looking at unseen landmark triplets. Since unseen triplets have no stand-alone estimation, we can propagate data from seen triplets with unary factors through the graph and examine how much information was propagated to the unseen triplets. Intuitively we expect the amount of propagated information to be dependent on both the initial amount of information in the graph and the topology of the unseen triplets relating to all of the seen triplets with unary factors.

The way we address this study is as follows: 
\begin{itemize}
	\item We suggest a simple \emph{information decay} model for describing composition-based propagation. This model was empirically derived also using a basic understanding of the process (described in section \ref{sec:Composition}).
	\item We suggest a simple algorithm for propagating data from seen triplets to unseen triplets by composition factors. Then we measure the amount of information in each variable node given its qualitative state probability using a unique \emph{information score} metric. 
	\item We use the information decay model to predict the information propagated to each unseen node. We call this prediction \emph{"topology Score"}. We show a reasonable correlation to the composition-based information score, which means we have a simple model for understanding how composition-based propagation depends on graph topology.
\end{itemize}

This study uses a simulation that generates random realizations of factor graphs that contain seen and unseen triplets with various levels of information and topologies as detailed in sections \ref{sec:Simulation} and  \ref{sec:FactorGraphPropagationAnalysis}. This section specifies our information decay model, graph propagation algorithm, information score, and topology score.

We use a few additional notations to describe our information decay model and propagation algorithm. Our algorithm needs to keep track of which triplet estimations have been updated in different stages of its operation. Therefore we divide the variable nodes $V_v$ into two subgroups: the set of triplets marked as updated $V_v^{u}$, and the set of triplets marked as not yet updated $V_v^{\neg u}$. The algorithm updates these groups during its operation.
\begin{equation*}
	V_v=V_v^{u}\cup V_v^{\neg u}.
\end{equation*}
For the reader's convenience, a recap of all factor graph-related notations can be found in Table \ref{table:graph_notations}.

\begin{table}[h!]
	\centering
	\scriptsize
	\begin{tabular}{ |p{0.1\textwidth}||p{0.7\textwidth}|  }
		\hline
		\multicolumn{2}{|c|}{qualitative factor graph notations} \\
		\hline
		notation & description \\
		\hline
		G & qualitative factor graph \\
		V & graph nodes \\
		E & graph edges \\
		$V_v$ & variable graph nodes (each corresponds to a triplet qualitative state $S^{i,j:k}$) \\
		$V_f$ & factor graph nodes \\
		$V_v^{seen}$ & variable graph nodes that where observed \\
		$V_v^{unseen}$ & variable graph nodes that where never observed together\\	
		$V_v^{u}$ & variable graph nodes that have already been updated during some stage of data propagation through the graph \\
		$V_v^{\neg u}$ & variable graph nodes that have not yet been updated during some stage of data propagation through the graph\\
		\hline
	\end{tabular}	
	\caption{Qualitative factor graph notations used in \ref{sec:FactorGraphPropagation}. }
	\label{table:graph_notations}
\end{table}

\vspace{5pt}
\subsubsection{Information Score} \label{sec:informationScorel}~\\
Our graph propagation algorithm and topology score refer to the information in each variable node. To measure the information of a triplet with a specific qualitative state probability $\prob{S^{AB:C}}$ (ground truth, estimated or prior), we introduce a new \emph{"information Score"} metric denoted as $ISC^{AB:C}$. To make this metric intuitive, we want it to maintain $0\le ISC^{AB:C} \le1$ where 0 means no information (uniform probability) and 1 means full information (qualitative state is perfectly known).

Respectively  we define \emph{information Score} of an arbitrary landmark triplet AB:C as the normalized entropy distance of its qualitative state probability from uniform probability:
\begin{equation}\label{eq:information_score}
	ISC^{AB:C} \triangleq \frac{\mathcal{H}_{max}-\mathcal{H}^{AB:C}}{ \mathcal{H}_{max}- \mathcal{H}_{min}},
\end{equation}
where $\mathcal{H}^{AB:C}$ is the entropy of the given qualitative state probability, $\mathcal{H}_{max}$ is maximal entropy (no information), and  $\mathcal{H}_{min}$ is minimum entropy (full information). Remembering the dimension of the discrete qualitative state probability vector is $d$ (dependent on the specific qualitative space partition we use), we get:
\begin{equation*}
	\begin{split}
		& \!\begin{multlined}[t]
			\mathcal{H}^{AB:C}= -\sum_{i=1}^{d}{\prob{S^{AB:C}=i|H^{AB:C}_n} }\\  \log{\prob{S^{AB:C}=i|H^{AB:C}_n}} 
		\end{multlined}\\
		& \mathcal{H}_{max}=\log{(\frac{1}{d})} \\
		& \mathcal{H}_{min}=0.
	\end{split}
\end{equation*}

\vspace{5pt}
\subsubsection{Information Decay Model} \label{sec:informationDecayModel}~\

A significant part of this study is finding a simplified model for the composition behavior in factor graph data propagation. Using a basic understanding of composition and empiric trial and error, we achieved two important insights: First, the qualitative composition cannot fully preserve the information and makes it decay (partially explained in section \ref{sec:Composition}). Second, the composition is a product of two problems (partially described in section \ref{sec:Composition}). Remembering that composition factors connected three triplet nodes (e.g. AB:C, BC:D, AB:D), the decay effect for the case where one triplet has information and the other two do not is roughly quadratic in reference to the case where two triplets have information and one does not. 

\begin{figure}[]
	\centering
	\subfloat[]{\includegraphics[width=0.4\textwidth]{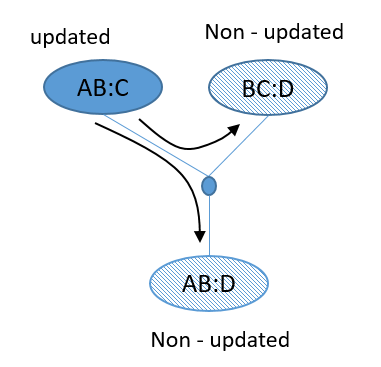}\label{fig:annihilation1} }\quad
	\subfloat[]{\includegraphics[width=0.4\textwidth]{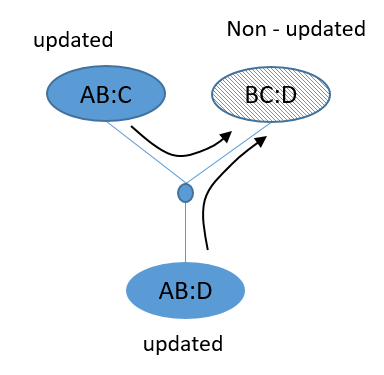}\label{fig:annihilation2} }\\
	\caption{\label{fig:InformationDecayModel} Each composition factor is connected to three variable nodes. Data is propagated through these factors during map update. Our information decay model for a single composition factor is dividend into two cases: (a) In case two variable nodes are non updated, and one is updated, decay model is $(1-\alpha)$  (b) I case one variable node is not updated and two nodes are, decay model is $(1-\alpha^2)$. (see section \ref{sec:informationDecayModel}). note that $0<\alpha<1$, so less information decay in the second case.}
\end{figure}

Therefore, we divide the model into two cases (see Figure \ref{fig:InformationDecayModel}). In case of two non-updated nodes and one updated node (e.g., $v_v^{BC:D}, v_v^{AB:D} \in V_v^{\neg u}; v_v^{AB:C} \in V_v^{u}$) we model the information score decays by a factor of $(1-\alpha)$ where $0<\alpha<1$:
\begin{equation}\label{eq:decay1}
	ISC^{BC:D}=ISC^{AB:D}= (1-\alpha) ISC^{AB:C},
\end{equation}
and in case one node is not updated, and two nodes are (e.g., $v_v^{AB:C}, v_v^{BC:D} \in V_v^{u}; v_v^{AB:D} \in V_v^{\neg u}$) we model a smaller decay factor of $(1-\alpha^2)$:
\begin{equation}\label{eq:decay2}
	ISC^{AB:D}= (1-\alpha^2) \frac{ISC^{AB:C}+ISC^{BC:D}}{2}.
\end{equation}
The scalar $\alpha$ is the decay factor and is empirically set to $\alpha=0.5$. Section \ref{sec:FactorGraphPropagationAnalysis} presents the results of this study and the correlation between "topological score" to actual ISC.

\vspace{5pt}
\subsubsection{Graph Propagation Algorithm} \label{sec:graphPropagationAlgorithm}~\

In the scope of this study, we consider factor graphs that contain seen and unseen triplets. We want to propagate the initial data from the seen triplets to the unseen ones using composition factors to test propagation by the composition process.

\begin{figure}[]
	\centering
	\subfloat[]{\includegraphics[width=0.45\textwidth]{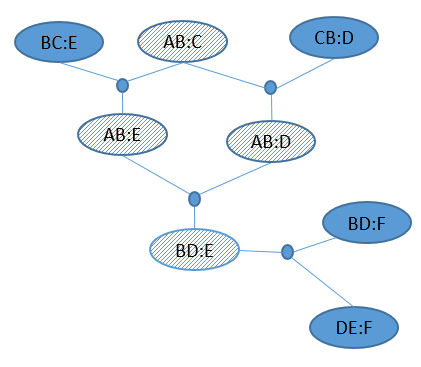}\label{fig:annihilation1} }\quad
	\subfloat[]{\includegraphics[width=0.45\textwidth]{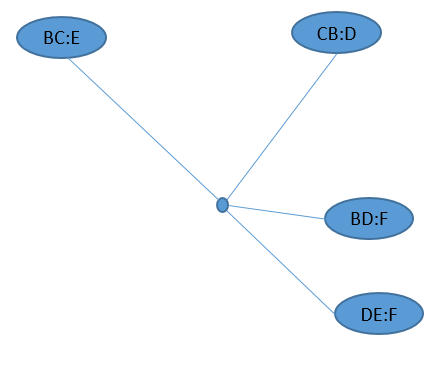}\label{fig:annihilation2} }\\
	\caption{\label{fig:factorGgraphElimination} Discrete factor graph variable elimination:  (a) a discrete qualitative factor graph contains only trinary (and unary) factors. If each discrete variable dimension id $d$, each trinary factor is a $d\times d\times d$ tensor. (b) eliminating variables $V_v^{AB:C}, V_v^{AB:D}, V_v^{AB:E}, V_v^{BD:E}$, we get a new factor that is a function of four qualitative states, and therefore is a $d\times d\times d\times d$ tensor. Similarly, multiple variables elimination may lead to factors that are a function of multiple variables and, therefore, are high dimensional tensors that are infeasible to compute.}
\end{figure}

The accurate way of doing this is by variable elimination (as discussed in \cite{dellart2006jrr}). Unfortunately, this approach is impractical in our case. Variable elimination replaces a group of connected variables and factor nodes with a single factor that connects all involved variables (see figure \ref{fig:factorGgraphElimination}). Generally, a discrete factor is a $d$ dimensional tensor with $d^{\# variables}$ entries (In our case, $d$ is the number of sections in the qualitative space partition). The factors generated in elimination usually involve multiple variables, are too big, and are practically infeasible in our case. Practically even very efficient implementations, such as in GTSAM \cite{gtsam} discrete factor graph, do not scale well.

Another well-known approximated solution is belief propagation, also known as sum-product message passing (see, e.g., \cite{su2015convergence}, \cite{braunstein2005survey}). This solution mainly holds for non-loopy graphs (which is not the case for us). Loopy belief propagation algorithms support general factor graph topology but are not proven to converge \cite{murphy2013loopy}.

For our study, we chose a simpler algorithm that is very stable and good enough to investigate the behavior of qualitative composition concerning graph topology. Finding a generally good approximated algorithm for this problem is still to be studied, and we leave this for future work. The algorithm we suggest inspired by label correcting algorithms is detailed in Alg.~\ref{alg:alg2}.

To better describe our algorithm, we add a few notations. During its progress, our algorithm needs to determine which composition factors are available for data propagation in each step. Accordingly, we divide the group of all composition factors $V_{fc}$ into two groups. The first group comprises composition factors connected to at least one updated triplet and one not updated triplet. These factors are considered available for propagating new information. We call these \emph{open} composition factors and denote this subgroup as $V_{fc}^{open}$. The second group includes composition factors that are connected only to updated or only to not updated triplets. These factors are considered unavailable for new information propagation and are called \emph{closed} composition factors. We denote this subgroup as $V_{fc}^{closed}$. The algorithm initializes both subgroups when it starts and updates them during its progress.
\begin{equation*}
	\begin{split}
		& V_{fc}=V_{fc}^{open}\cup V_{fc}^{closed}\\
		& V_{fc}^{open}= \{ v_{fc}^{ijks}\};\exists e(v_{fc}^{ijks}, v\in V_v^{u}) \land \exists e(v_{fc}^{ijks}, v\in V_v^{ \neg u})\\
		& V_{fc}^{closed}= \{ v_{fc}^{ijks}\};\nexists e(v_{fc}^{ijks}, v\in V_v^{u}) \lor \nexists e(v_{fc}^{ijks}, v\in V_v^{\neg u}).
	\end{split}
\end{equation*}

%
\begin{algorithm}
	\caption{qualitative factor graph data propagation}\label{alg:alg2}
	\footnotesize
	\begin{algorithmic}[1]
		\State {\color{blue} // init all seen triplets as updated}
		\State  $V_v^{u} = V_v^{seen}$
		\State  $V_v^{nu} = V_v \setminus V_v^{seen}$
		\State  init  $V_{fc}^{open},\, V_{fc}^{closed}$ by $V_v^{u},\, V_v^{\neg u}$
		\State  init  $ISC^{i,j:k} \forall\; v_v^{i,j:k} \in V_v^{seen}$
		\While {$V_v^{\neg u}$ not empty and $V_{fc}^{open}$ not empty}
		\State {\color{blue} // propagate all open composition factors}
		\ForAll{$v_{fc}^{i} \in V_{fc}^{open}$}     
		\State calculate qualitative state probability of non updated triplets -eq.~\eqref{eq:marginalComposition}
		\State  calculate ISC of non updated triplets
		\EndFor\label{euclidendwhile}
		\State {\color{blue} // keep only best ISC triplet}
		\State Find best new ICS triplet $v_v^{best} \in V_v^{\neg u}$
		\State mark best triplet as updated: $v_v^{best} \to V_v^{u}$
		\State keep $v_v^{best}$ estimation
		\State mark relevant composition factor as closed $V_{fc}^{i}\to V_{fc}^{closed}$
		\EndWhile
		\State {\color{blue} // handle unreachable triplets}
		\If{$V_v^{\neg u}$ \; \text{not empty} }
		\State $\forall \; v_v^{ij:k} \in V_v^{\neg u}$ set uniform state probability	
		\EndIf 			
	\end{algorithmic}
\end{algorithm}

\begin{figure}[]
	\centering
	\subfloat[]{\includegraphics[width=0.5\textwidth]{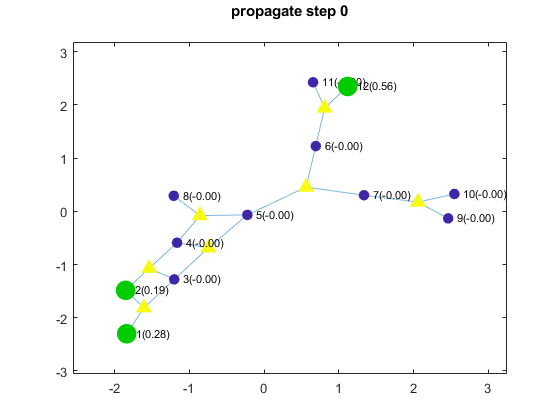}\label{fig:fgprp0} }\hspace{-22pt}\quad
	\subfloat[]{\includegraphics[width=0.5\textwidth]{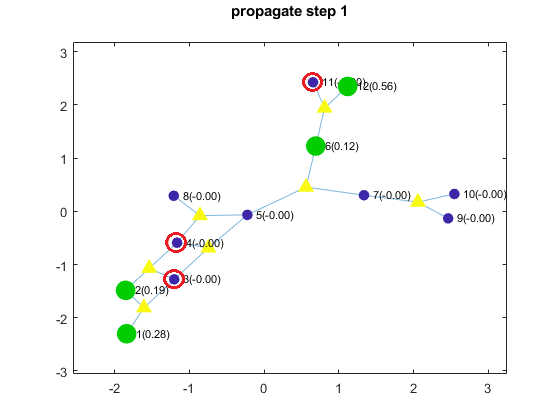}\label{fig:fgprp1} }\quad	
	\subfloat[]{\includegraphics[width=0.5\textwidth]{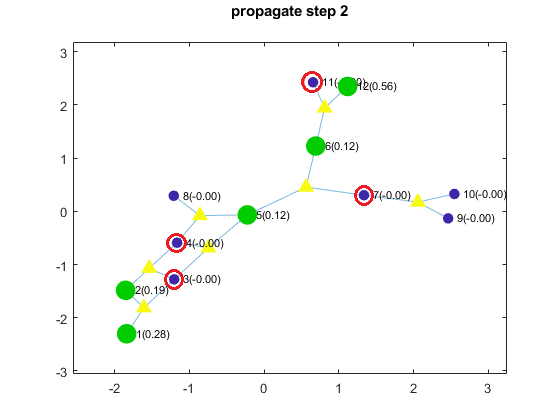}\label{fig:fgprp2} }\hspace{-22pt}\quad
	\subfloat[]{\includegraphics[width=0.5\textwidth]{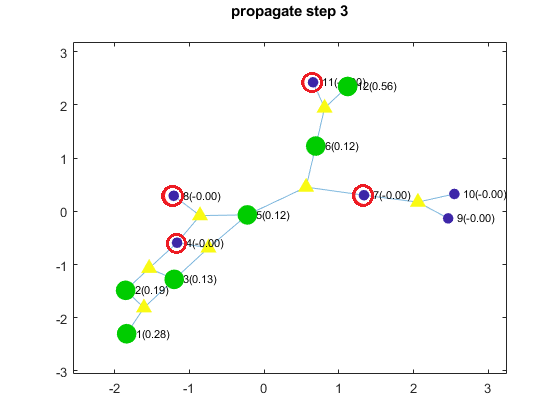}\label{fig:fgprp3} }\quad
	\caption{\label{fig:fgpropagation} Example for the factor graph propagation algorithm (described in \ref{alg:alg2}). In this example, we show a specific qualitative factor graph. Yellow triangles are composition factors. Small blue circles are non-updated variable nodes, while big green circles are updated nodes. Each variable node has a notation of Id(ISC level). In each step of the algorithm, several non-updated candidate nodes are considered for propagation (marked in red circles), but only the one with the best ISC is updated. (a) Initial state. Nodes 1,2,12 are source nodes and are marked as updated. (b) Nodes 3,4,6,11 are candidates to be updated. Node 6 gets the best ISC and is updated. (c) Nodes 3,4,5,7,11 are candidates to be updated. Node 5 gets the best ISC and is updated. (d) Nodes 3,4,7,8,11 are candidates to be updated. Node 3 gets the best ISC and is updated. (Remark: This example does not reach the end of the propagation). }
\end{figure}

In Alg.~\ref{alg:alg2} all triplets with unary factors $V_v^{seen}$ are "source" nodes. They are initialized as $V_v^{u}$ and are not updated again (lines 2-5). Every open composition factor connects updated and non-updated variable nodes. On every step of the algorithm, all open composition factors $V_{fc}^{open}$ propagate new data to their \emph{non-updated} triplets (lines 8-11). However, only the update of the factor that achieves the best information score for its \emph{not updated} nodes is saved. The selected triplet nodes are marked as \emph{updated} and are not updated again (lines 12-16). This way, every unseen node is updated only once. Finally, the algorithm halts when there are no open factors left. Figure \ref{fig:fgpropagation} shows an example of the graph propagation algorithm.

We chose this algorithm for three reasons: First, it is fast -  a single update for each triplet. Speedy runtime lets us go over many scenarios quickly and conduct a comprehensive study. Second, it is very stable and not affected by graph loops. Finally, it proved helpful for our purpose (as shown in section \ref{sec:FactorGraphPropagationAnalysis}).


\vspace{5pt}
\subsubsection{Topology Score} \label{sec:topologyScore}~\

The last part of this study shows how our information decay model explains the behavior of composition-based propagation in qualitative factor graphs. Our information decay model uses only the initial information of seen triplets (not the entire probability distribution) and graph topology to predict how much information will propagate to each variable node in the graph. We call this prediction \emph" topology score" (TSC) and denote it as $TSC(v_v^{AB:C})=TSC^{AB:C}$. In section \ref{sec:FactorGraphPropagationAnalysis}, we show that this prediction has a reasonable correlation to the ISC of unseen nodes from the composition-based propagation. Therefore we conclude it is a valid model.

The decay model prediction might be used when addressing a globally optimized solution for qualitative factor graph propagation or active qualitative planning.

The process of generating a "topology score" is as follows: we start by setting the topology score for all seen triplets $V_v^{seen}$ to be their initially estimated ISC:
\begin{equation*}
	TSC^{AB:C} = ISC^{AB:C} \quad \forall v_v^{AB:C} \in V_v^{seen}.
\end{equation*}
Then we use the same graph propagation algorithm described in Alg.~\ref{alg:alg2}, but instead of using actual composition factors for propagating data, we use our "information decay" model. The "topology score" variant of the algorithm is described in Alg.~\ref{alg:alg3}. Note that while we use Alg.~\ref{alg:alg2} to propagate actual data in the graph, our decay-based "topology score" can be applied to any graph propagation algorithm.
\begin{algorithm}
	\caption{topology score propagation}\label{alg:alg3}
	\footnotesize
	\begin{algorithmic}[1]
		\State {\color{red} \(\triangleright\) differences from in Alg.~\ref{alg:alg2} in red}
		\State {\color{blue} // init all seen triplets as updated}
		\State  $V_v^{u} = V_v^{seen}$
		\State  $V_v^{nu} = V_v \setminus V_v^{seen}$
		\State  init  $V_{fc}^{open},\, V_{fc}^{closed}$ by $V_v^{u},\, V_v^{\neg u}$
		\State  {\color{red} init  $TSC^{i,j:k}=ISC^{i,j:k} \quad \forall\, v_v^{i,j:k} \in V_v^{seen}$}
		\While {$V_v^{\neg u}$ not empty and $V_{fc}^{open}$ not empty}
		\State {\color{blue} // propagate all open composition factors}
		\ForAll{$v_{fc}^{i} \in V_{fc}^{open}$}     
		\State {\color{red} calculate TSC of non updated triplets - eqs.~\eqref{eq:decay1} and \eqref{eq:decay2}}
		\EndFor\label{euclidendwhile}
		\State {\color{blue} // keep only best {\color{red} TCS} triplet}
		\State Find best new {\color{red} TCS} triplet $v_v^{best} \in V_v^{\neg u}$
		\State mark best triplet as updated: $v_v^{best} \to V_v^{u}$
		\State keep $v_v^{best}$ estimation
		\State mark relevant composition factor as closed $v_{fc}^{i}\to V_{fc}^{closed}$
		\EndWhile
		\State {\color{blue} // handle unreachable triplets}
		\If{$V_v^{\neg u}$ \; \text{not empty} }
		\State $\forall \; v_v^{ij:k} \in V_v^{\neg u}$ set uniform state probability	
		\EndIf 		
	\end{algorithmic}
\end{algorithm}

Section \ref{sec:FactorGraphPropagationAnalysis} presents the results of this study and the correlation between "topological score" to actual ISC.

%% file: 06-Results.tex
We thoroughly evaluate our approach to test its performance and effectiveness. We discuss our test and analysis methodology in sections \ref{sec:Simulation} and  \ref{sec:PerformanceMetrics}. The first part of our approach concerns single landmark triplet estimation. In section \ref{sec:SingleTripletResults}, we use simulation to compare our approach to the state of the art \cite{Padgett17ras} and specifically show the effect of in-cooperating a motion model on speed and performance. We also compare our full, accurate algorithm to our approximated one (As presented in sections \ref{sec:-DetailedAlgorithm}, \ref{sec:faster variant}),  further improving the speed with a slight performance cost. In section \ref{sec:MRCLAMDatasetResults}, we use the  MCRLAM dataset \cite{Leung11ijrr} for testing in a more realistic scenario. 

The second part of our work addresses data propagation in the qualitative map. This part is innovative and without existing work as a reference. In section \ref{sec:BasicCompositionFactorAnalysis}, we evaluate the basic performance of qualitative composition, and in section \ref{sec:FactorGraphPropagationAnalysis}, we show the results of our study of composition behavior in a qualitative factor graph as discussed in section \ref{sec:FactorGraphPropagation}.

\subsection{Simulation}
\label{sec:Simulation}
\input{06a-Simulation}

\subsection{Performance Metrics}
\label{sec:PerformanceMetrics}
\input{06b-PerformanceMetrics}

\subsection{Single Triplet}
\label{sec:SingleTripletResults}
\input{06c-SingleTripletResults}

\subsection{Qualitative Map Data Propagation Results}
\label{sec:QualitativeMapDataPropagationResults}
\input{06bb-QualitativeMapDataPropagationResults}

%% file: 06a-Simulation.tex
\begin{figure}[]
	\centering
	\includegraphics[width=0.5\textwidth]{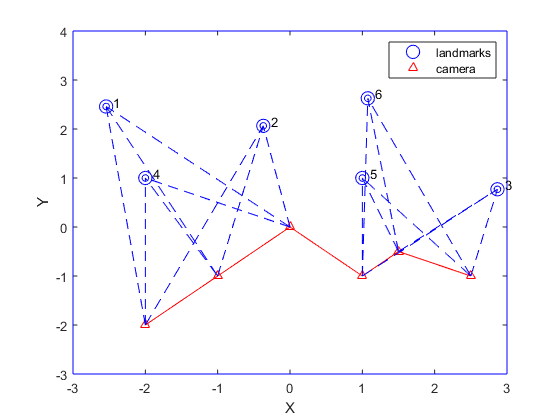}\label{fig:res12} 
	\caption{ \label{fig:scenario}Simplified illustration of a simulation scenario. The camera trajectory is presented in red. The camera gets measurements of two landmark triplets: 1,2:4 and 3,5:6. Each is observed from 3 different viewpoints. In the actual simulation, the order of observations is random.}
\end{figure}

The main tool we use to evaluate our approach is a MATLAB-based simulation. We consider a 2D scenario with point landmarks and a mobile camera. Specifically we uniformly choose positions for $m$ landmarks $L^i=(x_i,y_i), i\in1,...,m$. The set of landmarks is our metric ground truth map $M_{metric}=\{L^i\}_{i=1...m}$. Then we uniformly choose camera trajectory $Trajectory=\{X_j\}_{j=1...n}$ (where $X_j$ is camera pose at time instant $j$). We randomly select which landmarks are observed from each pose and randomize measurements and actions $H_n$ using Gaussian measurement and motion models. Camera to landmark measurements are noisy azimuth measurements, and actions are heading commands as described in Section \ref{sec:problem-formulation}. To evaluate the different parts of our work, we randomize numerous scenarios with various noise levels. Figure \ref{fig:scenario} illustrates an example scenario.

To keep the analysis general, we consider locations and poses as uniform as possible:  Landmark locations and camera trajectory are sampled uniformly; each pose observes only three landmarks. This constraint prevents biasing results with specific assumptions on camera trajectory or landmark visibility. Note that our method is more comprehensive than observing only three landmarks at a specific time. If a specific view observes more than three landmarks, all landmark triplets involved will generate measurements.

An important parameter of the simulation is its geometric dynamic range (\emph{GDR}), i.e.~the ratio between the minimal distance between any two objects and the maximal distance between any two objects (cameras or landmarks). All 2D locations are randomised within $x\in[-3,3]$ and $y\in[-3,4]$, and we enforce a minimum distance of $0.01$ between any two objects. This means $GDR\leq 800$. It is also a valid assumption for real systems (e.g., a robot moving in a $60$m $\times 70$m  area with landmarks and cameras not closer than $10$ cm from each other). Limiting the GDR enables a numerically stable solution.

We also note that landmark recognition is ideal with no false landmark identification. Our work focuses on basic qualitative geometry. Identification errors can be modeled in our approach (an important direction for future work).

%% file: 06b-PerformanceMetrics.tex
We use various metrics to analyze our inference performance. Using various metrics is an important addition compared to the analysis made in state-of-the-art approaches \cite{Mcclelland14jais} and \cite{Padgett17ras}. It provides us with a deeper understanding of qualitative inference. We use two main types of metrics:

\begin{enumerate}
	
	\item Ground truth (GT) related metrics:
	\begin{itemize}
		\item \emph{Probability DMSE:} The mean square error for the difference between the estimated qualitative state probability vector to the GT state vector $DMSE = \sqrt{\sum_{i=1}^{m}{(s_i^{AB:C}-s_{GT}^{AB:C})^2}}$. It tests how accurate the estimation is.
		\item \emph{Ground truth rating:} The rank of the GT qualitative state when ordering qualitative states by their estimation likelihood (1 is most likely).
		\item \emph{Geometric distance:} The mean distance from qualitative state centroids to GT state centroids weighted by the state estimated probability. $gmd=\sum_{i=1}^{m}{||c_i^{AB:C}-c_{GT}^{AB:C} ||_2}$ This metric tells us how close are the estimated states to GT (1 is the distance between A and B).
		\item\emph{Ground truth likelihood:} estimated posterior probability of GT qualitative state $\prob{s_{GT}^{AB:C}}$. This metric measures the accuracy of estimation but ignores false qualitative states.
		\item \emph{Ground truth likelihood ratio:} The ratio between the estimated probability of GT qualitative state to the estimated probability of the most likely state $\frac{\prob{s_{GT}}}{max(S^{AB:C})}$. This metric measures how close the GT state is to being most likely.
	\end{itemize}
	
	\item Information (or entropy) related metrics:
	\begin{itemize}
		\item \emph{Entropy:} estimation probability entropy $E = -\sum_{i=1}^m{\prob{s_i^{AB:C} \log{s_i^{AB:C}} } }$ measures how distributed is the qualitative state probability (or how much information is in the distribution). 
		\item \emph{Likelihood ratio:} the ratio between the second most likely qualitative state to the most likely qualitative state. This measures how "decisive" the estimation result is. Very important to maximal likelihood approaches.
	\end{itemize}
	
\end{enumerate}

Using all these metrics in different stages of result analysis teaches us much about the quality of our algorithms.

%% file: 06c-SingleTripletResults.tex
To evaluate the effect of our motion-model-incorporated inference, we compare three different algorithms: (a) State-of-the-art baseline \cite{Padgett17ras} (b) Our full multi-view inference as described in Section \ref{sec:Probabilistic navigation and mapping}. (c) Our fast approximated multi-view inference (section \ref{sec:Probabilistic navigation and mapping}).

The baseline for evaluating our performance is the previous work of \cite{Padgett17ras} (mapping only). To compare running time and complexity, we do not directly implement \cite{Padgett17ras}, but an equivalent algorithm. We use our formulation from Section \ref{sec:Probabilistic navigation and mapping} that leads to equations \ref{eq:slam1} and \ref{eq:slamst3}. Since \cite{Padgett17ras} does not consider a motion model, views are independent. Under these assumptions, we now get:
\begin{equation}\label{eq:slamInd1}
	\!\begin{multlined}[t]
		\prob{s_i^C | H_n} = \iint\displaylimits_{L^C \in s_i^C, X_{1:n}}{ \prob{X_{1:n},L^C| H_n}  dL^CdX_{1:n}},
	\end{multlined}
\end{equation}	
where
\begin{equation}\label{eq:slamInd2}
	\!\begin{multlined}[t]
		\prob{X_{1:n},L_C|H_n} = \frac{\prob{X_1,L_C}}{\prob{Z_1}}\prod_{i=1}^n{\frac{1}{\zeta_i}  \prob{Z_i|X_i,L_C} },
	\end{multlined}
\end{equation}	
and $\zeta_i \doteq \prob{Z_i|a_{1:i-1},Z_{1:i-1}}$. We then correspondingly implement a reduced version of our sample-based algorithm \ref{alg:alg1} in section \ref{sec:-DetailedAlgorithm}.

To evaluate the single triplet inference, we randomize 300 scenarios with 36 different combinations of measurement and motion model noise levels for each scenario (10800 scenarios in total). Measurement and motion model has Gaussian noise as described in section \ref{sec:problem-formulation}, where measurement noise is $\sigma_v \in [0^{\circ},10^{\circ}]$ and motion model noise is $\sigma_w \in [0^{\circ},20^{\circ}]$. We also use landmark triplets with views from 3-time steps, as shorter trajectories will not show the effect of using a motion model, and longer trajectories typically do not improve results significantly.

\vspace{15pt}
\subsubsection{General performance}
\begin{figure}[]
	\centering
	\subfloat[]{\includegraphics[width=0.50\textwidth]{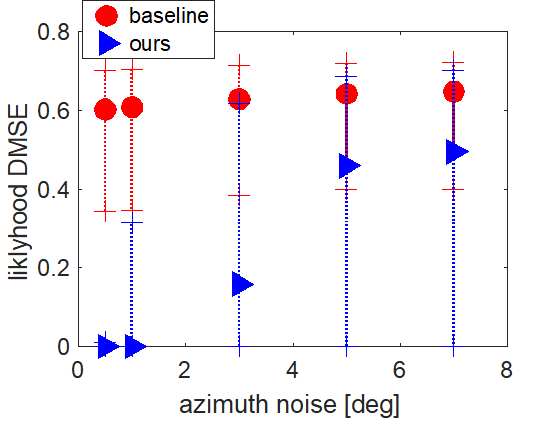}\label{fig:res11} }\hspace{-20pt}\quad
	\subfloat[]{\includegraphics[width=0.50\textwidth]{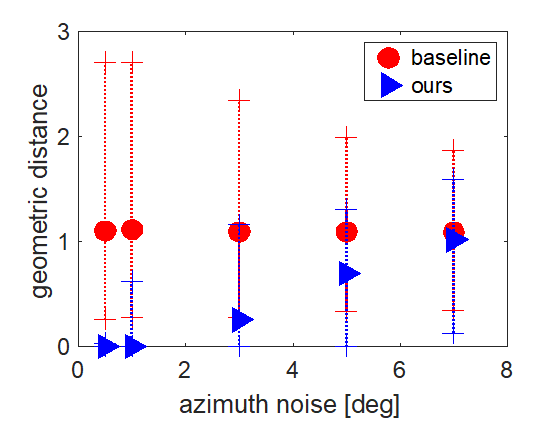}\label{fig:res12} }\quad	
	\caption{\label{fig:results1} EDC single triplet estimation. Our motion model in-cooperated algorithm is compared to the state-of-the-art baseline \cite{Padgett17ras}. The plot shows median and percentiles 25 and 75 for each algorithm. (a) DMSE Vs. measurement noise (motion model noise is $\times 2$). (b) mean geometric distance vs.~measurement noise (motion model noise is $\times 2$).}
\end{figure}

~\\
 Figure \ref{fig:results1} shows performance results and compares our motion-model-incorporated inference to the state-of-the-art. Performance is represented by \emph{DMSE} and \emph{Geometric distance} as specified in section  \ref{sec:PerformanceMetrics}. In this table, for the ease of visualization, results are a function of azimuth measurement noise, while motion model heading noise is $\times2$ bigger for each run correspondingly.

\begin{table}[h!]
	\centering
	\scriptsize
	\begin{tabular}{ |p{0.22\textwidth}||p{0.18\textwidth}|p{0.18\textwidth}|p{0.18\textwidth}|  }
		\hline
		\multicolumn{4}{|c|}{EDC estimation with motion model} \\
		\hline
		Metric& baseline & ours & ours-fast\\
		\hline
		DMSE & 0.39, 0.63, 0.71 & \textbf{0, 0.16, 0.63} & 0, 0.21, 0.62\\
		geometric distance & 0.28, 1.10, 2.30 & \textbf{0, 0.25, 1.15} & 0, 0.27, 1.16\\
		Entropy & 0.28, 0.66, 0.87 & \textbf{0, 5e-3, 0.58} & 0, 0.07, 0.64\\
		time[sec] & 26 & 18 & \textbf{0.05}\\
		\hline
	\end{tabular}
	\caption{EDC state estimation for single triplet. measurement noise: $\sigma_v=2^\circ$, motion model noise: $\sigma_w=5^\circ$}.
	\label{table:1}
\end{table}
We observe that using a motion model dramatically improves performance. Up to measurement noise of $2^\circ$ (which is reasonable even for low-end camera-based platforms), results are much better than state-of-the-art and almost perfect. Up to $7^\circ$ (which is a significant error for camera-based systems), our estimated probability \emph{DMSE} is still better than state-of-the-art. We can also see that for $7^\circ$ $\emph{gmd}<1$, all of the qualitative states that are wrongly estimated are likely to be close to the ground truth state.

Another significant result concerns our fast-approximated algorithm. Table \ref{table:1} shows various performance metrics, including run time for each algorithm. We show 25 percentile, median, and 75 percentile. Our fast approximated algorithm achieves performance very close to the full one. While our full algorithm is about $\times2$ faster than the baseline, the approximation is about $\times100$ faster than both. These results show a successful usage of the ability of the coarse qualitative estimation to absorb errors and enable fast approximations. While we use a simple MATLAB implementation, absolute run times are irrelevant but can be used for comparing the algorithms.

\begin{figure}[]
	\centering
	\subfloat[]{\includegraphics[width=0.46\textwidth]{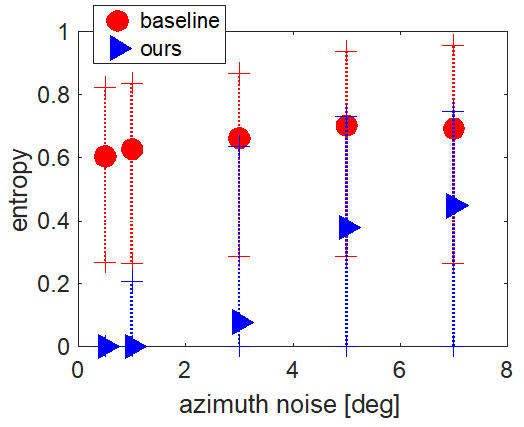}\label{fig:res21} }\hspace{-15pt}\quad
	\subfloat[]{\includegraphics[width=0.50\textwidth]{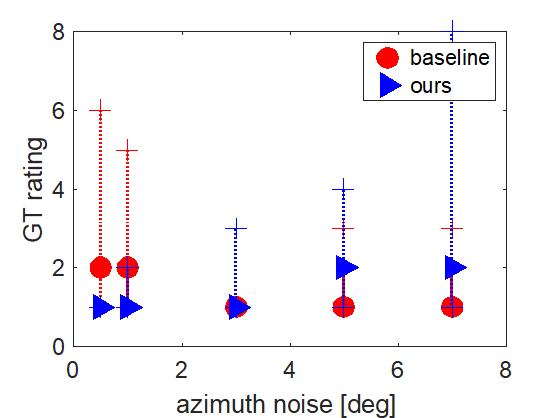}\label{fig:res22} }\quad	
	\caption{\label{fig:results2} EDC single triplet estimation. Our motion model in-cooperated algorithm is compared to the state-of-the-art baseline \cite{Padgett17ras}. The plot shows median and percentiles 25 and 75 for each algorithm. (a) Entropy Vs. measurement noise (motion model noise is $\times 2$). (b) Ground truth rating vs.~measurement noise (motion model noise is $\times 2$).}
\end{figure}

Figure \ref{fig:results2} includes additional performance metrics. Unlike \emph{DMSE} and \emph{Geometric distance}, which measure the correctness of the estimation, \emph{Entropy} measures only the steepness of it. Entropy performance shows similar behavior. On the other hand, looking at \emph{ground truth rating}, we note that for high noise levels, our estimation is sometimes lesser than the baseline method. Remembering that even in these conditions, the correctness of our probability distribution is better (as seen in \emph{DMSE}) we conclude that this is caused by the fact that scenarios with higher noise may lead to multiple (mostly two and up to 4) likely qualitative states. In these cases, minor estimation errors may translate to a change in likelihood rating between these close states. We also note that these high noise levels are unlikely for real camera-based applications.

\vspace{15pt}
\subsubsection{Fast Variant Noise Sensitivity}~\\
Observing that the fast approximation algorithm presents results close to the full algorithm, we analyzed its performance more extensively. Since our fast variant is noise ignorant (as detailed in section \ref{sec:faster variant}), we want to specifically test the different effects of motion model noise and measurement noise separately. We use the same simulation and scenario set but look at different cross-sections of noise levels.

\begin{figure}[]
	\centering
	\subfloat[]{\includegraphics[width=0.50\textwidth]{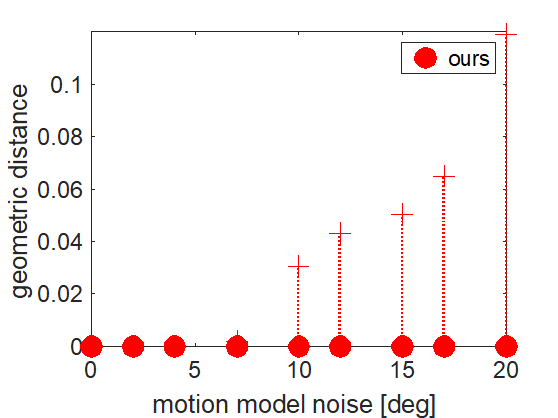}\label{fig:mm_geodist} }\hspace{-20pt}\quad
	\subfloat[]{\includegraphics[width=0.50\textwidth]{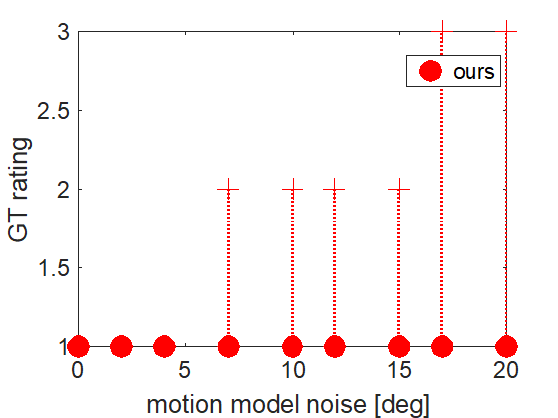}\label{fig:mm_GTrating} }\quad
	\subfloat[]{\includegraphics[width=0.50\textwidth]{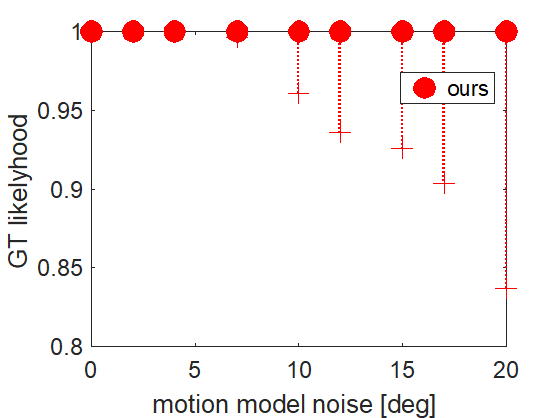}\label{fig:mm_GTliklihood} }\\	
	\caption{\label{fig:mmResultsl} Sensitivity to motion model error.  The plot shows median and percentiles 25 and 75. (a) mean geometric distance Vs. Motion model noise. (b) ground truth rating Vs. Motion model noise (c) ground truth likelihood Vs. Motion model noise.}
\end{figure}

Figure \ref{fig:mmResultsl} shows the effect of motion model noise on the performance. We look at scenarios with 0 measurement noise and various motion model noise levels. Motion model noise has a small effect on ground truth likelihood and geometric distance. Even with a very high noise of $20^\circ$, the estimated probabilities are relatively unaffected.  However, it has more effect on the ground truth rating (but only on the 75 percentile), although  GT is typically in the top 3 most likely states.

We conclude that this is caused by scenarios with more challenging geometry (landmarks close to qualitative state edges or ill-conditioned camera poses) that may lead to two or three likely qualitative states. In these cases, the minor errors may translate to a change in likelihood rating between these close states.

\begin{figure}[]
	\centering
	\subfloat[]{\includegraphics[width=0.50\textwidth]{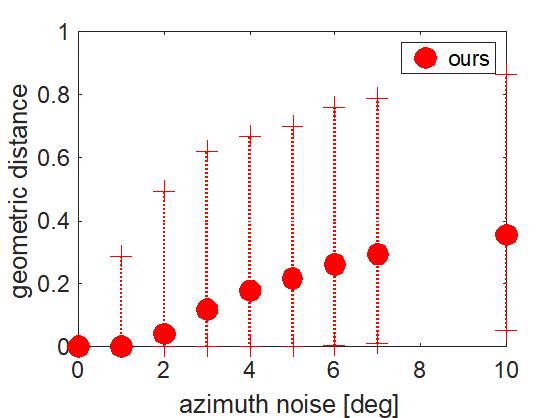}\label{fig:az_geodist} }\hspace{-20pt}\quad
	\subfloat[]{\includegraphics[width=0.50\textwidth]{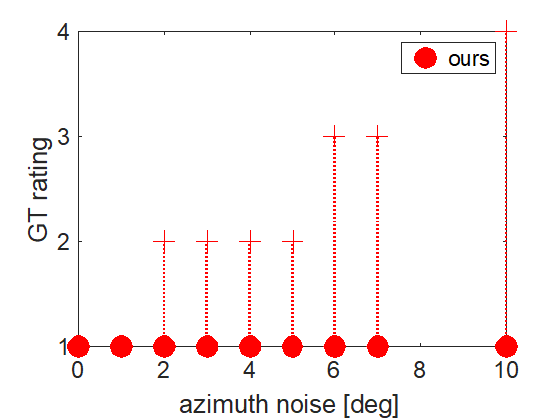}\label{fig:az_GTrating} }\quad
	\subfloat[]{\includegraphics[width=0.50\textwidth]{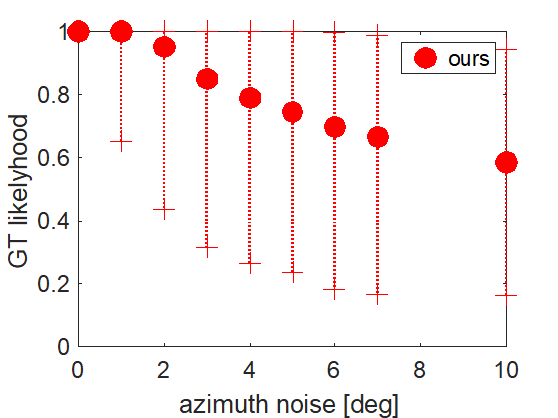}\label{fig:az_GTliklihood} }\\	
	\caption{\label{fig:azResultsl}  Sensitivity to measurement model error. The plot shows median and percentiles 25 and 75. (a) mean geometric distance Vs. Measurement model noise. (b) ground truth rating Vs. Measurement model noise (c) ground truth likelihood Vs. Measurement model noise. }
\end{figure}

Figure \ref{fig:azResultsl} shows the effect of measurement noise on the performance. We look at scenarios with 0 motion mode noise and various azimuth measurement noise levels. The effect of measurement noise is more substantial. Both geometric distance and ground truth likelihood respond as noise increases above $2^\circ$, but still, medians are not very strongly affected. Again GT rating responds harder on the 75 percentile but is stable on the median, leading us to the same conclusions.

\vspace{15pt}
\subsubsection{MRCLAM Dataset Results}~\\
\label{sec:MRCLAMDatasetResults}
\input{06d-MRCLAMDatasetResults}

\vspace{15pt}
\subsubsection{Discussion}~\\
Generally, our full algorithm and fast approximation have several good properties. First, they perform very well for azimuth measurement noise of up to $2^\circ$ and motion model heading noise of up to $10^\circ$, which is a reasonable range even for low-quality camera-based platforms. Secondly, they are more sensitive to measurement model noise than motion model noise, which is also practical since measurement noise is usually smaller in real-world scenarios. Furthermore, performance degradation in high noise levels is gradual and generally stable. 

We also successfully demonstrate the unique attributes of qualitative inference to hold approximations well and achieve a very low compute global nonlinear algorithm. We believe there is a good potential for this approach to be practical for real-world qualitative autonomous navigation and mapping. It may also even be more meaningful when addressing online qualitative active planning.

%% file: 06d-MRCLAMDatasetResults.tex
To evaluate our method in a realistic scenario, we use the \emph{MRCLAM} dataset \cite{Leung11ijrr}, which comprises several scenarios of multiple robots moving around pre-set landmarks with unique markers. We choose this dataset since it has robot pose GT and landmark identification and location GT. In addition, landmarks are round and, therefore, act as point landmarks. This real-world dataset allows us to evaluate how informative our qualitative estimation framework is (e.g.~in terms of entropy) and quantify performance concerning GT, which has yet to be tackled in previous work \cite{Padgett17ras}. 

We use five scenarios, each with five robots and 15 landmarks. We get 16-230 landmark triplets observed three times or more in each scenario. Our qualitative approach thus generates rich mapping for these scenarios.

\begin{figure}[]
	\centering
	\includegraphics[width=0.7\textwidth]{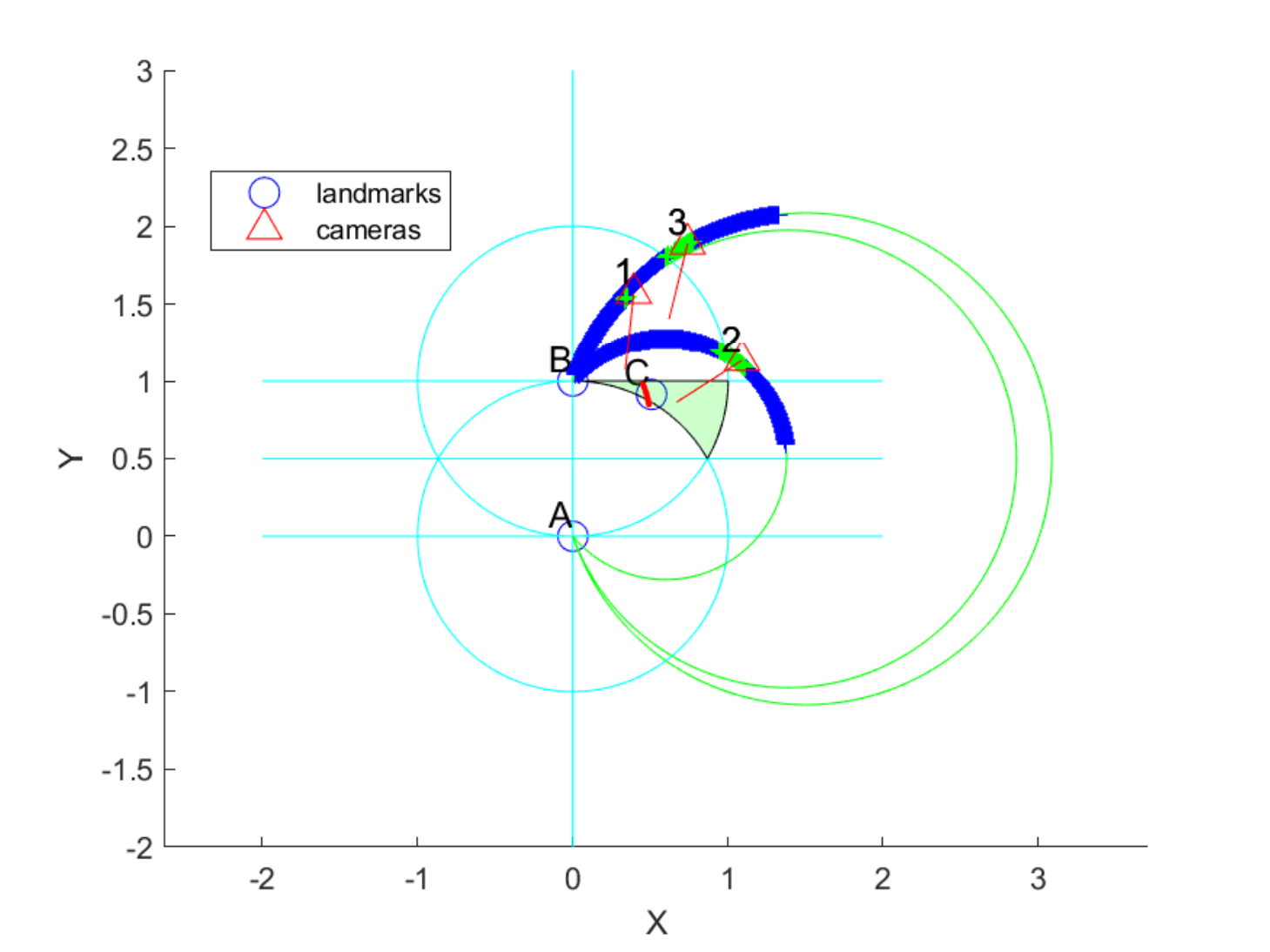} 
	\caption{\label{fig:res_example}\label{fig:MRCLAM} MRCLAM dataset: EDC example scenario result. Ground truth camera poses are red triangles, and ground truth landmark locations are blue circles. Thin green lines are camera locus circles, while thick blue and green are valid hypotheses after two and three view constraints. Redline is the landmark C location hypothesis after three-view constraints. All estimations are close to GT, and the error is still inside GT's qualitative state (light green area).}
\end{figure}

A typical estimation result for a single landmark triplet observed from three camera views is in Figure \ref{fig:res_example}. We can see that the usage of the motion model reduces the trajectory sampling hypotheses $TH$ (blue color), which is further refined by the landmark $C$ triangulation step to only a small subset (green color). The resulting landmark $C$ location hypotheses are in red. One can observe that the camera and landmark hypotheses are very close to GT. As a result, the likelihood of the GT qualitative state in this example run is 1.

Table \ref{table:2} summarizes the results for this dataset. We compare our fast algorithm variant to an uninformative uniform distribution. Looking at \emph{Entropy} and \emph{DMSE}, we conclude that estimation is informative and is close to the truth. \emph{GT rating} shows that most of the time, the GT state is the most likely, and always within the top 2. Also, looking at \emph{gmd}, we can see that all likely qualitative states are close to GT. In conclusion, we note that results show meaningful and informative estimation, which concurs with our stimulative results considering the reasonable noise levels in the dataset.

\begin{table}[h!]
	\centering
	\scriptsize
	\begin{tabular}{ |p{0.13\textwidth}||p{0.20\textwidth}|p{0.13\textwidth}|  }
		\hline
		\multicolumn{3}{|c|}{MRCLAM dataset EDC estimation} \\
		\hline
		& ours-fast & uniform \\
		\hline
		DMSE & 0.03, 0.45, 0.69 & 0.97\\
		gmd & 5e-3, 0.27, 0.71 & 2.2\\
		Entropy &4e-3, 0.38, 0.69 & 3\\
		GT rating &   1,    1,    2 & -\\
		\hline
	\end{tabular}	
	\caption{MRCLAM dataset results summary: We show 25 percentile, median, and 75 percentile for each metric. Our approximated fast algorithm is compared to an uninformative uniform estimation. }
	\label{table:2}
\end{table}

%% file: 06bb-QualitativeMapDataPropagationResults.tex
We now analyse data propagation in the qualitative map as described in section \ref{sec:ApproachComposition}. First, in section \ref{sec:BasicCompositionFactorAnalysis} we show the results of a basic single triplet simulation that measures how much information is propagated through a single trinary composition factor (as described in section \ref{sec:Composition}). Then, in section \ref{sec:FactorGraphPropagationAnalysis} we present the results of our qualitative graph data propagation study (as described in section \ref{sec:FactorGraphPropagation}). We discuss the specifics of how we simulate the problem, and show the correlation between our topological information decay model to actual composition composition based propagation.

\vspace{15pt}
\subsubsection{Basic Composition Factor Analysis}~\\
\label{sec:BasicCompositionFactorAnalysis}
\input{06e-BasicCompositionFactorAnalysis}

\vspace{15pt}
\subsubsection{Factor Graph Propagation Analysis}~\\
\label{sec:FactorGraphPropagationAnalysis}
\input{06f-FactorGraphPropagationAnalysis}

\vspace{15pt}
\subsubsection{Discussion}~\\
\label{sec:MapPropagationConclusion}
Looking at data propagation in the qualitative map gives us several important insights. First, the qualitative composition can propagate significant information, even to landmark triplets that were never seen together.

We also show that our information decay model reasonably predicts how much information is propagated by composition. Hence, we have a good simple model for understanding data propagation in the qualitative map regarding its topology and existing information. 

This process of data propagation and the innovative connection between data propagation and graph topology are essential tools when addressing real-world qualitative autonomous planning.

%% file: 06e-BasicCompositionFactorAnalysis.tex
In this section, we evaluate the basic composition operator for propagating data between three common landmark triplets as detailed in section  \ref{sec:Composition}. While composition can improve existing estimation, we only test its performance in propagating data to un-estimated triplets, enabling us to understand its operation clearly.

Specifically, we simulate 1000 scenarios, each comprising three inter-connected landmark triplets $(\text{AB:C, BC:D, AB:D})$ and camera trajectories with different noise levels. Histories, in terms of measurements and controls $H^{AB:C}$ and $H^{BC:D}$ are available, but $H^{AB:D}=\emptyset$. We estimate $\prob{S^{AB:C}|H^{AB:C}}$ and $\prob{S^{BC:D}|H^{BC:D}}$ using measurement history and our fast approximated algorithm. Then, $\prob{S^{AB:D}|H^{AB:C}, H^{BC:D}}$ is inferred only through composition.
\begin{figure}[]
	\centering
	\subfloat[]{\includegraphics[width=0.50\textwidth]{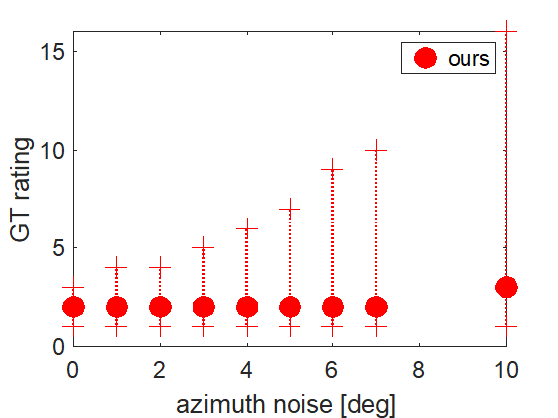}\label{fig:res21} }\hspace{-20pt}\quad
	\subfloat[]{\includegraphics[width=0.50\textwidth]{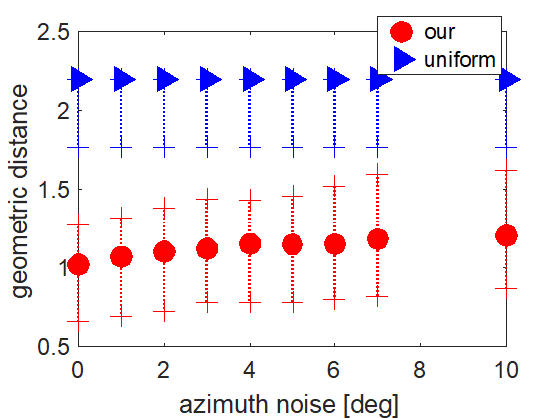}\label{fig:res22} }\quad	
	\subfloat[]{\includegraphics[width=0.50\textwidth]{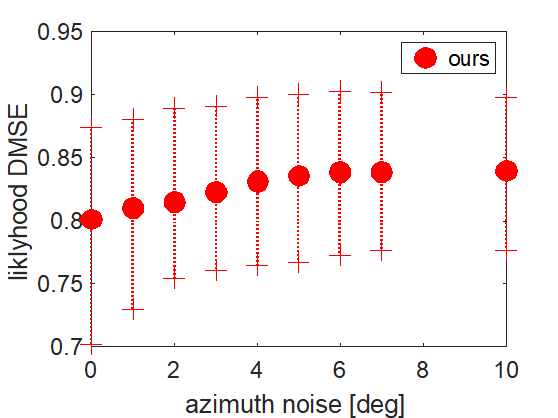}\label{fig:res23} }
	\caption{\label{fig:results_composition1} EDC probabilistic composition. The plot shows median and percentiles 25 and 75. (a) DMSE vs.~measurement noise. Composition results Vs. uniform probability (qualitative state unknown) (b) mean geometric distance vs.~measurement noise. In addition to our composition, we also show results for uninformative estimation. }
\end{figure}
Figure \ref{fig:results_composition1} summarizes the performance evaluation. The propagated data is compared to the initial state of uniform estimation for the unseen triplets. Results are not perfect, but considering this triplet has no direct measurements, we see that composition still propagates significant information. The GT rating tells us that for reasonable measurement noise of up to $2^{\circ}$, the GT state is mostly among the four most probable EDC states (and half the time among the two leaders). The $gmd$ plot shows that the False EDC states are mainly close to the GT state.
Considering that these are triplets that were never measured together and could not be estimated otherwise, these are significant results that might enable the use of unseen triplets in active qualitative planning.



%% file: 06f-FactorGraphPropagationAnalysis.tex
The last part of our work studies how composition factors propagate data through the qualitative factor graph. As explained in section \ref{sec:FactorGraphPropagation} this is done in 4 steps:
\begin{enumerate}
	\item We generate a factor graph with seen triplets that has an initial independent estimation and unseen triplets with no independent estimation.
	\item We use the composition-based Alg. \ref{alg:alg2} to propagate data from seen triplets to all unseen triplets and calculate the "information score" for each one (section \ref{sec:graphPropagationAlgorithm}).
	\item We use our information decay model and Alg. \ref{alg:alg3} to generate a "topology score" for all triplets  (sections \ref{sec:informationDecayModel}, \ref{sec:topologyScore}). This score is a prediction of the amount of information that will get to each triplet based on our decay model and graph topology.
	\item Finally, we compare the "topology score" to the actual "information score" for each triplet and see if the information decay model correlates to the actual composition-based propagation.
\end{enumerate}

Next, we specify this study in more detail.

\vspace{5pt}

\subsubsection{Factor Graph Simulation}~\\
First, we must simulate scenarios with complex factor graphs containing seen and unseen triplets in various topologies. To do so, we use the same simulation described in section \ref{sec:Simulation} to organize landmarks and camera poses uniformly and randomly. Then we consider all possible landmark triplets $V_v^{full}$, and all possible composition factors $V_{fc}^{full}$. We choose a random subgroup of $n_{fc}$ composition factors $V_{fc} \subseteq V_{fc}^{full}$, and all related triplets $V_v \subseteq V_v^{full}$ to make the factor graph. Then out of all the $n_v=\left|V_v\right|$ triplets we randomly choose $n_{seen}=\frac{n_v}{\text{coverage rate}}$ triplets to have observations. Coverage rate is a parameter of the simulation that affects how close unseen triplets will be to the closest seen triplet. A high coverage rate means most triplets are seen, so many unseen triplets are directly connected to seen triplets via composition. A low coverage rate means few triplets are seen, so many unseen triplets are a few composition factors away from the closest seen triplet. We empirically set it to 0.5, which gives us a good variety of typologies. Azimuth and motion model measurements are then randomized using the proper noise models (section \ref{sec:problem-formulation}) from camera trajectory to the selected triplets. The rest are unseen triplets $V_v^{unseen}$.

\begin{figure}[]
	\centering
	\subfloat[]{\includegraphics[width=0.5\textwidth]{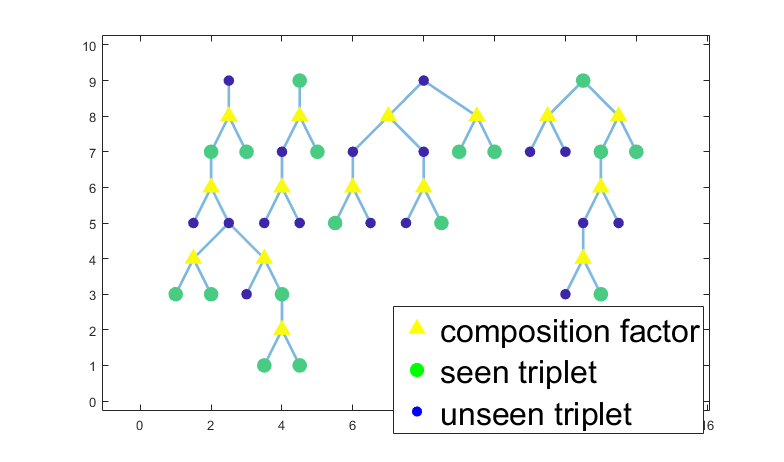}\label{fig:fgres11} }\hspace{-20pt}\quad
	\subfloat[]{\includegraphics[width=0.35\textwidth]{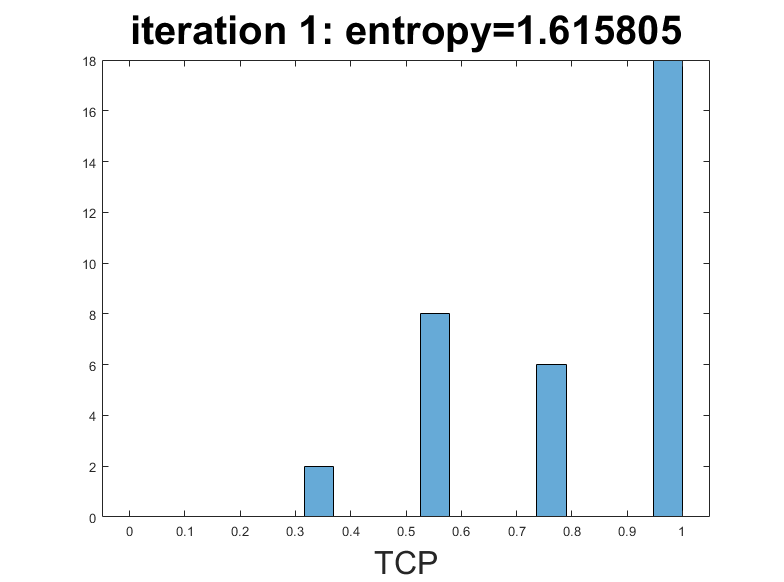}\label{fig:fgres12} }\vspace{-20pt}\quad	
	\subfloat[]{\includegraphics[width=0.5\textwidth]{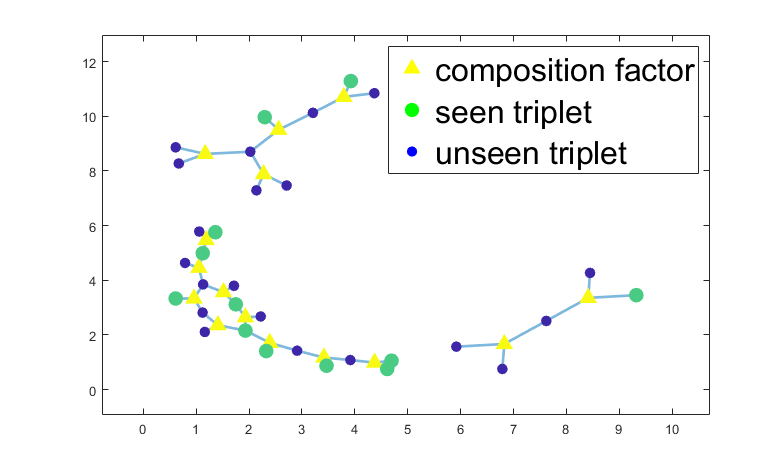}\label{fig:fgres21} }\hspace{-20pt}\quad
	\subfloat[]{\includegraphics[width=0.35\textwidth]{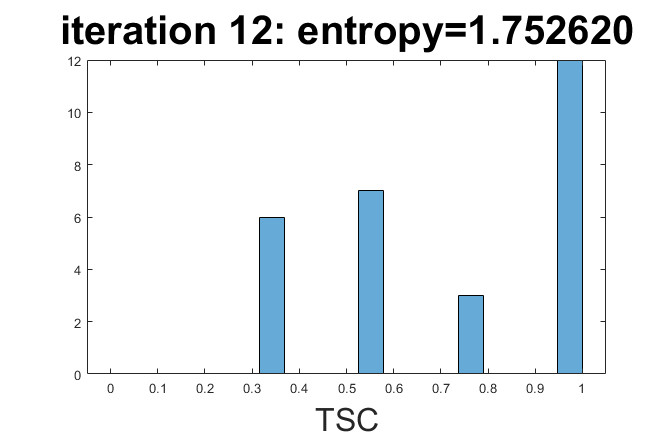}\label{fig:fgres22} } \vspace{-20pt}\quad
	\subfloat[]{\includegraphics[width=0.5\textwidth]{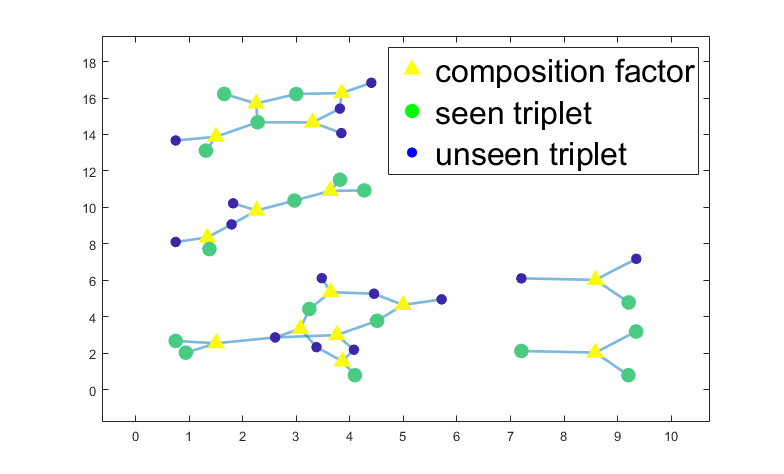}\label{fig:fgres31} }\hspace{-20pt}\quad
	\subfloat[]{\includegraphics[width=0.35\textwidth]{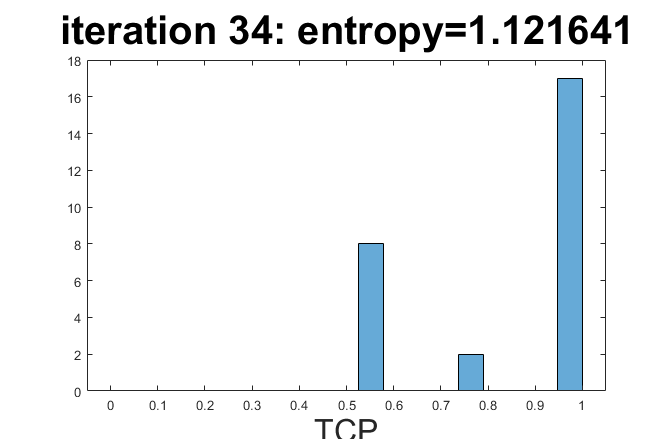}\label{fig:fgres32} } \vspace{-20pt}\quad
	\subfloat[]{\includegraphics[width=0.5\textwidth]{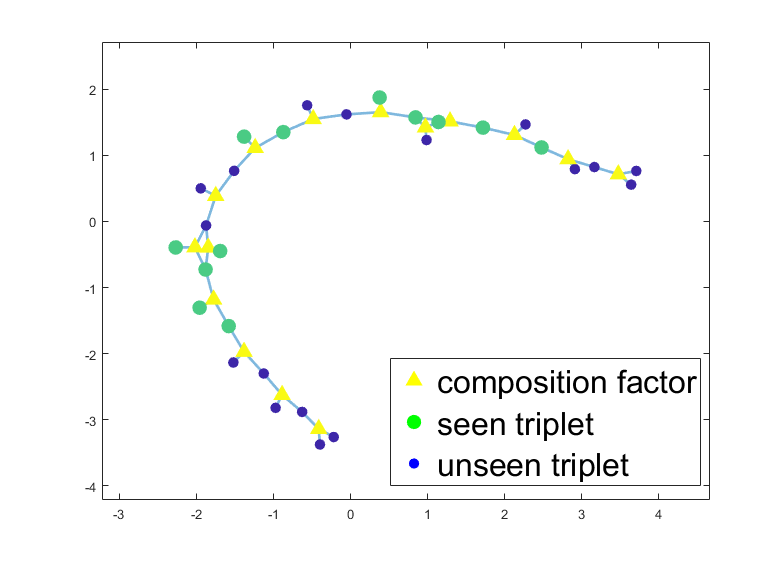}\label{fig:fgres41} }\hspace{-20pt}\quad
	\subfloat[]{\includegraphics[width=0.35\textwidth]{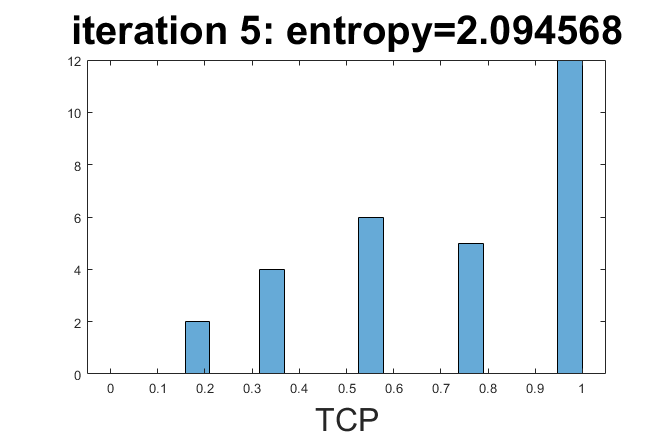}\label{fig:fgres42} } \quad
	\caption{\label{fig:fgSelection} For generating factor graph scenarios, we randomize multiple graphs and use a TSC histogram to select graphs with various connectivity levels. (a), (c), (e), and (g) are examples of selected factor graphs. (b), (d), (f), and (h) are the corresponding TPC histograms and connectivity scores. }
\end{figure}

For our study, we prefer factor graphs with various levels of connectivity between seen and unseen triplets. Therefore, we also apply a process for selecting factor graphs with various connectivity levels. Given a factor graph, we calculate TSC for each triplet as described in \ref{sec:topologyScore}. Since measurement-based estimation is not relevant in this stage, we just initialize TSC to be 1 for all seen triplets $TSC^{i,j:k}=1$, $ \forall V_v^{i,j:k} \in V_v^{seen}$ and 0 for all $V_v^{unseen}$, and we apply Alg. \ref{alg:alg3}. We look at the histogram of topology scores of all triplets in the graph and calculate the corresponding entropy (figure \ref{fig:fgSelection}). We call this \emph{connectivity score}. It is a measure of the variability of topology in the graph. If the connectivity score is high, it means the distribution of TCP in the graph is close to uniform, and a low connectivity score means there are only a few dominant TCP values in the graph. Therefore we randomly generate multiple-factor graphs and choose the best connectivity score. 

\vspace{5pt}
\subsubsection{Information Decay Model vs.~Composition}~\\
\label{sec:DecayVsComposition}
With 5400 simulated factor graph scenarios, we solve all seen triplets using their measurements and our fast variant of the single triplet solver (section \ref{sec:faster variant}). We then use the composition-based Alg. \ref{alg:alg2} to propagate data from seen triplets to all unseen triplets and calculate the information score for each one (section \ref{sec:graphPropagationAlgorithm}).

Then we generate a \emph{topology score} for all triplets with our information decay model and Alg.~\ref{alg:alg3} as detailed in sections \ref{sec:informationDecayModel} and  \ref{sec:topologyScore}. This score predicts the amount of information that will get to each triplet based on graph topology. Finally, we compare the topology score to the actual information score for each triplet. 

\begin{figure}[]
	\centering
	\includegraphics[width=0.7\textwidth]{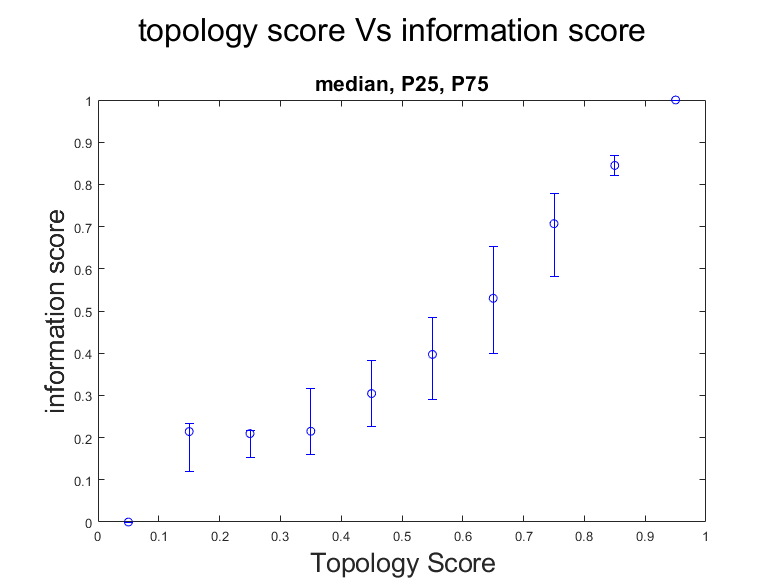} 
	\caption{\label{fig:fgres_7}\label{fig:decayModelCorrelation} The correlation between the Information decay model and composition. Topology scores of all triplets in all scenarios are collected into ten bins. For each bin, we show the percentile 25, median, and percentile 75 statistics of the corresponding triplets information scores.}
\end{figure}

Figure \ref{fig:decayModelCorrelation} shows how our information decay model prediction relates to the actual composition-based information score. In this figure, we collect topology scores of all triplets in all scenarios and bin them into ten bins. For each bin, we collect the information scores of the corresponding triplets and show percentile 25, median, and percentile 75 statistics. We see that the information decay model correlates reasonably to the composition-based propagation. 

This is a validation of our information decay model and the main contribution of this study. Having a simplified model for qualitative factor graph data propagation means we understand this process, which also might be valuable for active qualitative planning.

To test if our information decay model is trivial, we also tried several other simplified models for predicting qualitative data propagation via composition. An example is a model we call \emph{composition level} (CL). This model sets the composition level for seen triplets $V_v^{seen}$ to be 0. Propagation via a composition factor just increases the composition level by 1. We denote the composition level of a triplet node $v_v^{AB:C}$ to be: $CL^{AB:C}$. As in section \ref{sec:informationDecayModel} we address the propagation through a composition factor that involves three triplet nodes (e.g., AB:C, BC:D, AB:D) and divides the model into two cases. In case of two non-updated nodes and one updated node (e.g., $V_v^{BC:D}, V_v^{AB:D} \in V_v^{\neg u}; V_v^{AB:C} \in V_v^{u}$):
\begin{equation}\label{eq:cl1}
	CL^{BC:D}=CL^{AB:D}= CL^{AB:C}+1,
\end{equation}
and in case one node is not updated, and two nodes are (e.g., $V_v^{AB:C}, V_v^{BC:D} \in V_v^{u}; V_v^{AB:D} \in V_v^{\neg u}$):
\begin{equation}\label{eq:cl2}
	CL^{AB:D}= \min\{CL^{AB:C}, CL^{BC:D}\} + 1.
\end{equation}
Then composition level is propagated through the graph using an algorithm similar to Alg.~\ref{alg:alg3}.

\begin{figure}[]
	\centering
	\subfloat[]{\includegraphics[width=0.50\textwidth]{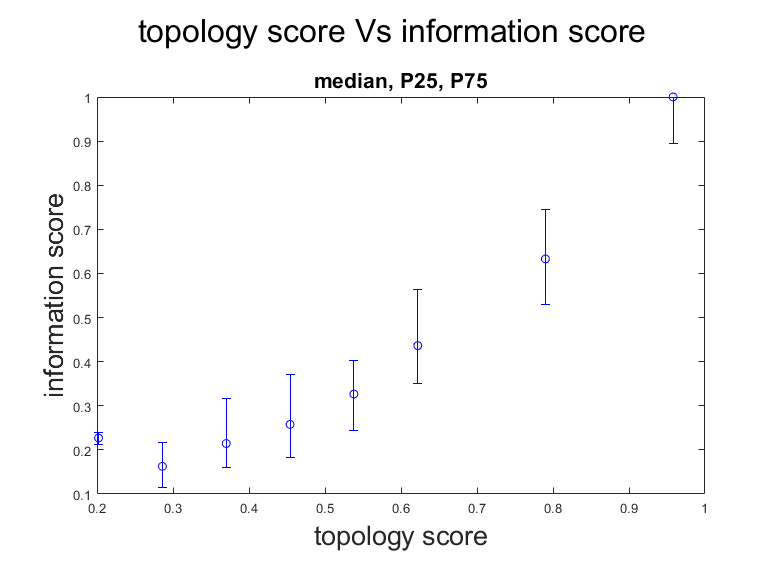}\label{fig:fgres_7} }\hspace{-20pt}\quad
	\subfloat[]{\includegraphics[width=0.50\textwidth]{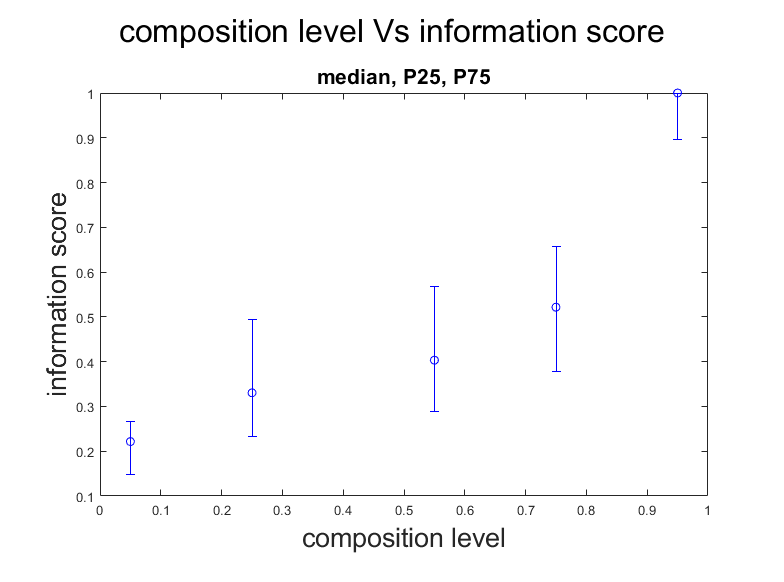}\label{fig:fgres_7} }\quad	
	\caption{\label{fig:comareModels} Comparing two simplified models for qualitative data propagation to actual composition-based information score (as described in \ref{sec:DecayVsComposition}). Topology score (a) represents our information decay model; (b) is an alternative composition level model. For both scores, all triplets from all scenarios are collected into (the same) 10 bins. For each bin, percentile 25, median, and 75 statistics are shown. (a) exhibits a better correlation than (b). }
\end{figure}

Comparing the information decay-based topological score to the composition level model, we see that TCP correlates more to the actual composition-based information score (figure \ref{fig:comareModels}). Since the composition level does not use the information in the unary factors, we use a simplified version of TCP to make an unbiased comparison, using the same information decay model and graph propagation algorithm but initializing TCP to be 1 for all  $V_v^{seen}$ instead of $ISC(\prob{S^{ij:K}})$. Statistics binning is done the same way for both. The composition level graph is incomplete because it naturally gets limited values but is binned into the same ten bins. Also, to make the graphs comparable, in figure \ref{fig:comareModels}, we present a normalized version of the composition level:

\begin{equation}\label{eq:cl3}
	\hat{CL}^{ij:k}= 1-\frac{CL^{ij:k}}{ CL_max},
\end{equation}
where $CL_max$ is the single maximal composition level in the graph.

We conclude that only some ways of modeling composition factor propagation are valuable and that the information decay model is noticeably useful.

%% file: 07-Conclusions.tex
This paper presents a new approach for localization and mapping based on qualitative spatial reasoning. We use a new and more natural and general formulation to achieve several innovations and improvements, taking a few steps towards a whole and valuable qualitative SLAM framework. We enable incorporating a motion model in qualitative estimation to improve run time and performance. Our method allows easy use of various qualitative space partitions and underlying SLAM solvers. To demonstrate the benefits of qualitative inference, we suggest a sampling-based global non-linear algorithm that does not need initialization and then develop a very fast approximated algorithm with minimal compromise on performance. 

Furthermore, we show how to represent the problem using a factor graph and propagate data efficiently using the inherent properties of qualitative geometry. We also infer landmark triplets with no direct observations, an essential building block for active qualitative planning. Furthermore, we give a simple model for understanding the behavior of this qualitative data propagation. Our approach can be complemented by finding more general or optimal algorithms for data propagation in the factor graph and understanding their performance limits.

In addition to improving complexity and performance compared to the state-of-the-art \cite{Mcclelland14jais} and \cite{Padgett17ras}, we show that our approach gets practical performance for low-compute mapping and localization in cases where the exact metric location is not essential. 

This work lays the basics for further investigating and expanding qualitative localization and mapping. We see an indication that additional research may lead to a practical low-compute alternative for low-cost autonomous agents in various tasks. Some of the directions we envision for future research are handling recognition errors, addressing complex volumed landmarks and 3D geometry, active qualitative planning \cite{Zilberman22iros}, and building a large and diverse qualitative SLAM or active planning dataset based on existing standard datasets.
